\documentclass[10pt,logo,copyright]{nvidiatechreport}
\usepackage[authoryear,round,sort]{natbib}
\let\cite\citep
\definecolor{citecolor}{HTML}{0071BC}
\definecolor{linkcolor}{HTML}{ED1C24}

\makeatletter
\@ifpackageloaded{hyperref}{}{
    \usepackage[pagebackref=false,breaklinks=true,colorlinks,citecolor=citecolor,linkcolor=linkcolor,bookmarks=false]{hyperref}
}
\makeatother

\hypersetup{
    pdftitle={Into the LidarVerse: A Real-Time Unified LiDAR World Model Across Tasks and Sensors},
    pdfauthor={Henry Che, Xuanchi Ren, Sanja Fidler, Shenlong Wang, Laura Leal-Taixe, and Shengyu Huang}
}

\usepackage{epsfig}
\usepackage{graphicx}
\usepackage{amsmath,mathtools}
\usepackage{amssymb}
\usepackage{amsthm}
\usepackage{nicefrac}       %
\usepackage{microtype}      %
\usepackage{amsmath}
\usepackage{float}
\usepackage{amsfonts}
\usepackage{bm}
\usepackage{enumitem}
\usepackage{mathtools}
\usepackage{multirow}
\usepackage{siunitx}
\usepackage{makecell}

\usepackage{fontawesome}
\usepackage{pifont}
\usepackage{color}
\usepackage[algo2e,inoutnumbered,linesnumbered,algoruled,vlined]{algorithm2e}
\usepackage{algpseudocode}
\usepackage{tabularx}
\usepackage{url}
\usepackage{bbm}
\usepackage{threeparttable}
\usepackage{tikz}
\usepackage{wrapfig}
\usepackage{placeins}
\usetikzlibrary{backgrounds}
\usetikzlibrary{decorations.pathreplacing}
\usepackage[outline]{contour}
\usepackage{caption}
\usepackage{subcaption}
\usepackage{wrapfig}
\usepackage{dashrule}
\usepackage{wrapfig}
\usepackage{booktabs}

\theoremstyle{definition}

\DeclarePairedDelimiterX{\infdivx}[2]{(}{)}{%
  #1\;\delimsize\|\;#2%
}

\definecolor{darkblue}{RGB}{49,130,189}
\definecolor{stanfordgrey}{RGB}{46,45,41}
\definecolor{cardinalred}{RGB}{253,141,60}

\makeatletter
\DeclareRobustCommand\onedot{\futurelet\@let@token\@onedot}
\def\@onedot{\ifx\@let@token.\else.\null\fi\xspace}

\makeatother

\usepackage[capitalize]{cleveref}

\crefname{section}{\S}{\S\S}
\crefname{subsection}{\S}{\S\S}
\crefname{conj}{Conj.}{Conj.}

\Crefname{assumption}{\textbf{H}\hspace{-3pt}}{\textbf{H}\hspace{-3pt}}
\crefname{assumption}{\textbf{H}}{\textbf{H}}

\crefname{algorithm}{Alg.}{Alg.}
\crefname{assumption}{\textbf{H}}{\textbf{H}}
\crefname{equation}{Eq.}{Eq.}
\crefname{definition}{Dfn.}{Dfn.}
\crefname{lemma}{Lemma}{Lemma}
\crefname{dfn}{Dfn.}{Dfn.}
\crefname{thm}{Thm.}{Thm.}
\crefname{tab}{Tab.}{Tab.}
\crefname{fig}{Fig.}{Fig.}
\crefname{table}{Tab.}{Tab.}
\crefname{figure}{Fig.}{Fig.}

\definecolor{mygreen}{RGB}{159, 200, 59}
\definecolor{myred}{RGB}{223, 135, 102}

\definecolor{policy_green}{rgb}{0.2, 0.59, 0.2}
\definecolor{policy_yellow}{rgb}{0.8, 0.8.0, 0.2}

\usepackage{amsmath,amsfonts,bm,mathtools}

\def\eqref#1{equation~\ref{#1}}

\def\1{\bm{1}}

\DeclareMathAlphabet{\mathsfit}{\encodingdefault}{\sfdefault}{m}{sl}
\SetMathAlphabet{\mathsfit}{bold}{\encodingdefault}{\sfdefault}{bx}{n}

\def\@onedot{\ifx\@let@token.\else.\null\fi\xspace}

\newcommand{\methodname}{LidarVerse\xspace}

\title{Into the LidarVerse: A Real-Time Unified LiDAR World Model Across Tasks and Sensors}
\author{Henry Che$^{1,2}$, Xuanchi Ren$^{3}$, Sanja Fidler$^{3}$, Shenlong Wang$^{2}$, Laura Leal-Taixé$^{1}$, Shengyu Huang$^{1}$ \\
$^1$ NVIDIA, $^2$ University of Illinois, Urbana-Champaign, $^3$ University of Toronto \\
\url{https://research.nvidia.com/labs/sil/projects/lidarverse}}

\begin{document}
\maketitle
\begin{figure}[h]
    \vspace{-2em}
    \centering
    \includegraphics[width=1\linewidth]{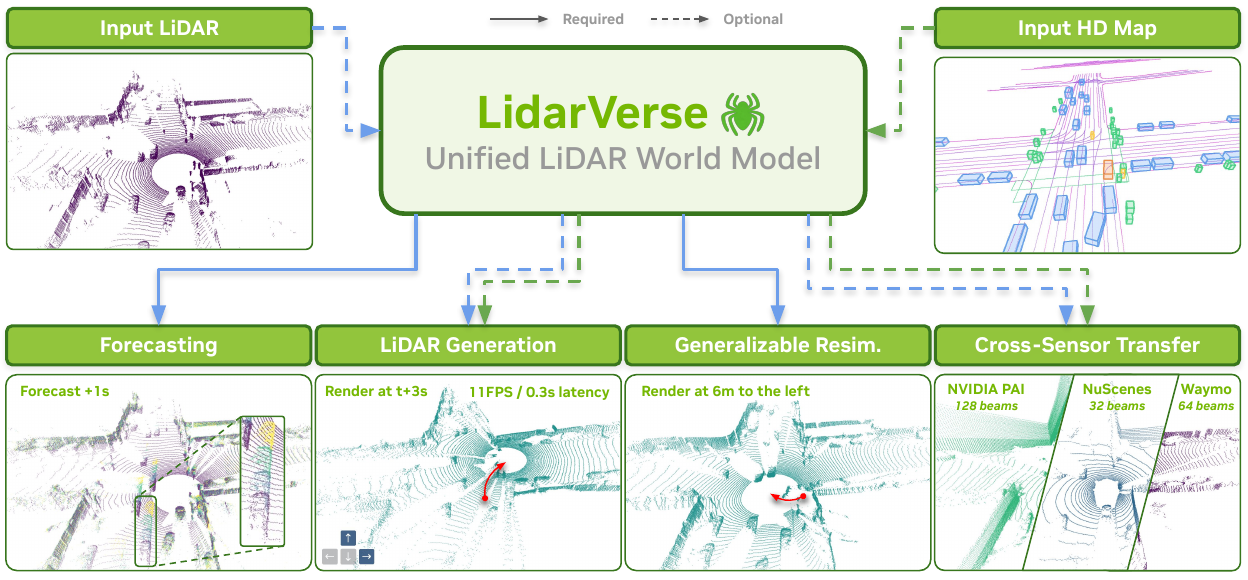}
    \caption{\textbf{\methodname} unifies generative and reconstruction-grounded LiDAR sensor simulation. By changing the available scene evidence, one distilled causal checkpoint generates new worlds, forecasts future point clouds, or resimulates observed scenes along new sensor trajectories with real-time, long-horizon inference.}
    \label{fig:teaser}
\end{figure}

\begin{abstract}
We present \methodname{}, a real-time unified LiDAR world model for sensor simulation across tasks and sensors. Existing LiDAR simulators largely follow one of two complementary paradigms: reconstruction-based methods preserve the geometry of a captured scene but require lengthy per-scene optimization, whereas generative methods amortize learning across data but lack the consistency needed for long-horizon rollouts. \methodname{} brings both paradigms into a single, unified conditional generative framework. A universal tokenizer maps LiDAR range sequences into a shared latent space, where a bidirectional teacher learns a common interface for conditioning on a LiDAR prefix, an HD-Map, a persistent spatial cache, and optionally, text. Varying the available scene evidence turns the same model into a generator of new worlds, a point-cloud forecaster, or a generalizable resimulator of observed scenes without per-scene optimization. We distill the teacher into a four-step autoregressive causal student and maintain both a rolling temporal K/V cache and a world-aligned spatial cache, enabling persistent long-horizon rollouts with real-time inference at 11\,FPS on a single H100 GPU. \methodname{} achieves state-of-the-art performance on most reported metrics across all three evaluation protocols, while its tokenizer supports 32-, 64-, and 128-beam LiDAR and generalizes to an unseen scanning pattern. These results demonstrate that generative and reconstruction-grounded LiDAR simulation can be realized within one model, providing the core capabilities for future closed-loop sensor simulation.
\end{abstract}

{\hbadness=10000\abscontent\par}

\tableofcontents
\newpage

\section{Introduction}
\label{sec:introduction}

Sensor simulation is critical for developing autonomous systems. Robots and self-driving vehicles must operate safely across a long tail of environments, trajectories, and interactions that cannot be collected exhaustively in the real world. Recent reconstruction-based systems such as NVIDIA Omniverse NuRec~\cite{nvidia2024nurec} transform captured driving logs into interactive simulation environments, while generative simulators such as OmniDreams~\cite{nvidia2026omnidreams} synthesize controllable observations beyond the captured data. Together, they highlight two complementary capabilities: preserving an observed world while changing the sensor trajectory, and creating new worlds beyond recorded experience. LiDAR is especially important in this setting because it provides direct geometric measurements under challenging illumination, yet its sparse, viewpoint-dependent returns are difficult to simulate faithfully. A practical LiDAR simulator should offer both capabilities while producing accurate observations, remaining temporally coherent under sensor motion, and running fast enough for interactive use.

Neural LiDAR simulation has largely developed along the same divide. \emph{Reconstruction-based} methods fit an explicit neural scene representation to a captured log and render LiDAR from novel sensor poses~\cite{dynfl,huang2023nfl,zheng2024lidar4d,zhou2025lidarrt}. Their persistent 3D representations provide strong geometric consistency and precise viewpoint control, making them well suited to scene resimulation. However, fitting each scene can require hours of optimization, and the resulting simulator remains tied to a manually collected log. These methods can vary the sensor trajectory within a scene, but do not readily scale to a diverse distribution of new environments.

\emph{Generative} methods instead amortize scene modeling across data and learn a distribution from which novel LiDAR scenes can be sampled~\cite{zyrianov2024lidardm,liu2025lagen,liu2025lidarcrafter}. This formulation promises diversity and scale, but realizing that promise depends on a suitable representation. Operating directly on unordered point sets is inefficient and prevents straightforward reuse of mature image and video generation infrastructure. Range images provide a structured, sensor-native alternative that preserves LiDAR scanning topology, but high-resolution range sequences are expensive to model directly with pixel-space diffusion~\cite{zyrianov2022lidargen,nakashima2023r2dm,nakashima2024r2flow}. Meanwhile, tokenizers developed for RGB video are poorly matched to range images, and existing LiDAR tokenizers generally target a single scanning pattern~\cite{ran2024lidm,hu2024rangeldm}. LiDAR generation therefore still lacks a compact latent interface that scales across both time and sensors. As a result, existing generators tend to specialize in one sensor configuration, produce only short sequences, or require many denoising steps. The field thus lacks a unified simulator that combines the distributional coverage of generative models with the scene fidelity, persistence, and control of reconstruction-based methods.

We address this gap with \methodname{}, a real-time unified LiDAR world model that brings generative and reconstruction-based simulation into one conditional framework. Its foundation is a universal, fully attention-based LiDAR tokenizer designed specifically for range images. Local spatial neighborhood attention captures sparse geometric structure, while factorized temporal attention exchanges information across sweeps without reducing the native temporal resolution. This design accommodates different scan patterns within a shared representation, allowing one tokenizer checkpoint to reconstruct data from 32-, 64-, and 128-beam sensors and generalize to an unseen scanning pattern. It also provides a common latent space in which heterogeneous LiDAR datasets can contribute to the same world model.

On this shared latent space, we train a bidirectional diffusion teacher with a unified interface for a LiDAR prefix, an HD-Map and actor layout, a persistent spatial cache (PSC), and, optionally, text. Rather than assigning each task a separate architecture or prediction head, we sample these conditions independently during training and use their composition to specify the inference regime. With no observed LiDAR geometry, the model generates new worlds from scene-level conditions; an observed LiDAR prefix turns the same model into a point-cloud forecaster; and a PSC built from observed sweeps anchors the model to a particular world for generalizable LiDAR resimulation under a new sensor trajectory without per-scene optimization. Generation, forecasting, and resimulation are therefore different ways of querying the same learned world, distinguished by the available evidence rather than by model parameters.

To make this unified model practical for interactive simulation, we distill the bidirectional teacher into a four-step block-autoregressive causal student. A rolling temporal K/V cache retains recent dynamics, while a world-aligned PSC preserves scene geometry beyond the finite temporal window, enabling persistent and temporally coherent long-horizon rollouts. The resulting causal checkpoint retains the shared conditioning interface and performs all three tasks at or above the native 10-Hz LiDAR rate. Across their respective evaluation protocols, \methodname{} outperforms task-specific baselines on most reported metrics. Together, these results establish a unified sensor-simulation paradigm in which changing the conditions, rather than the architecture, moves one model between generative simulation and reconstruction-grounded resimulation. To summarize, our main contributions are:
\Needspace{6\baselineskip}
\begin{itemize}[leftmargin=*]
    \item \textbf{Universal LiDAR tokenization.} We introduce a fully attention-based tokenizer that preserves native temporal resolution, improves geometric reconstruction and generative modeling, and supports 32-, 64-, and 128-beam sensor configurations within one checkpoint.
    \item \textbf{Unified LiDAR sensor simulation.} We bridge generative and reconstruction-based simulation within one conditional world model. By varying the available conditions, a shared checkpoint performs LiDAR generation, point-cloud forecasting, and generalizable LiDAR resimulation without task-specific heads or per-scene optimization.
    \item \textbf{Real-time, persistent world modeling.} We distill the bidirectional teacher into a few-step causal student and introduce a rolling PSC for consistent long-horizon generation. The resulting model operates in real time and achieves state-of-the-art results across the three evaluated task protocols.
\end{itemize}

\section{Related Work}
\label{sec:related-work}
Prior LiDAR sensor simulation methods typically specialize in one operating regime: generating novel scenes, forecasting future observations, or reconstructing a captured scene for resimulation. These regimes differ primarily in how much scene evidence is available at inference. \methodname{} spans this spectrum with one conditional world model: scene-level conditions guide LiDAR generation, an observed prefix enables forecasting, and a persistent spatial cache anchors generalizable resimulation without per-scene optimization. We review the task-specific literature below and identify the capabilities that the unified model inherits from each line of work.

\subsection{LiDAR Generation}
Early LiDAR generation methods focus on synthesizing individual sweeps in point, voxel, BEV, or range-image representations~\cite{zyrianov2022lidargen,xiong2023ultralidar, nakashima2023r2dm,nakashima2024r2flow,ran2024lidm,hu2024rangeldm}.
LiDARGen~\cite{zyrianov2022lidargen} pioneered diffusion-based range-image synthesis, establishing a strong foundation for data-driven LiDAR simulation. Subsequent methods extended synthesis beyond individual sweeps by generating temporally coherent sequences or 3D/4D geometry from which LiDAR observations can be rendered~\cite{zyrianov2024lidardm,li2025uniscene,ren2024scube,xu2026u4d}. Cosmos-Drive-Dreams~\cite{cosmosdrivedreams} generates 4D LiDAR using a single-frame tokenizer and relies on rich conditioning from HD-Maps and multiview cameras to ensure consistency.
More recently, LiDARCrafter~\cite{liu2025lidarcrafter} and LaGen~\cite{liu2025lagen} formulate LiDAR sequence generation autoregressively, while Sensor2Sensor~\cite{wang2026sensor2sensor} autoregressively converts monocular video into multimodal driving logs. However, these methods are limited to short rollouts and do not maintain consistency over long horizons.
In contrast, \methodname{} combines few-step self-forcing distillation with persistent 3D memory, maintaining coherent long rollouts while generating at the native sensor rate in real time.

\subsection{LiDAR Resimulation}
LiDAR resimulation reconstructs a captured scene and renders it along novel sensor trajectories~\cite{yang2023unisim,tonderski2024neurad,dynfl,tao2023lidarnerf,zheng2024lidar4d,zhou2025lidarrt,turki2026simuli}.
Most existing approaches rely on per-scene optimization, which requires lengthy reconstruction and tends to overfit the recorded trajectory, leading to degraded quality under large viewpoint shifts.
For camera resimulation, generalizable reconstruction methods reduce reconstruction cost~\cite{storm,chen2024g3r,lu2025drivingrecon,wang2025flux4d,chen2026dggt,nvidia2026instantnurec}, while another line of work incorporates generative priors to repair artifacts in under-observed regions~\cite{wu2025difix3d,che2026genre,schmid2026genrecon,yin2026gsfixer,zhu2026gaussfusion,wei2026gsfix3d}.
Both advances remain largely unexplored for LiDAR resimulation.
\methodname{} combines these benefits for generalizable LiDAR resimulation: a single cross-scene generative model requires no per-scene optimization, completes content beyond the captured trajectory, and renders native spinning-LiDAR observations in real time.

\subsection{Point-Cloud Forecasting}
Point-cloud forecasting predicts future sensor observations from a short
history.  4D-Occ~\cite{khurana2023point} pioneered the formulation of this task
as spacetime occupancy forecasting followed by LiDAR rendering.  UnO~\cite{agro2024uno}
learns a continuous occupancy field without labels, and DIO~\cite{diehl2025dio}
further decomposes occupancy and scene flow into individual instances.
Related approaches forecast point clouds from visual
observations~\cite{yang2024vidar} or model future scenes with discrete generative
occupancy~\cite{hu2024copilot4d}.  Most occupancy-based forecasters employ a
task-specific predictor for a prescribed future horizon.  \methodname{} instead
uses the same probabilistic LiDAR generator as its world model: forecasting is
obtained by conditioning on an observed prefix, without a discriminative
forecasting head, and autoregressive sampling can continue beyond the temporal
horizon seen during training.

\section{Method}
\label{sec:method}

\begin{figure}[!htbp]
    \centering
    \includegraphics[width=1\linewidth]{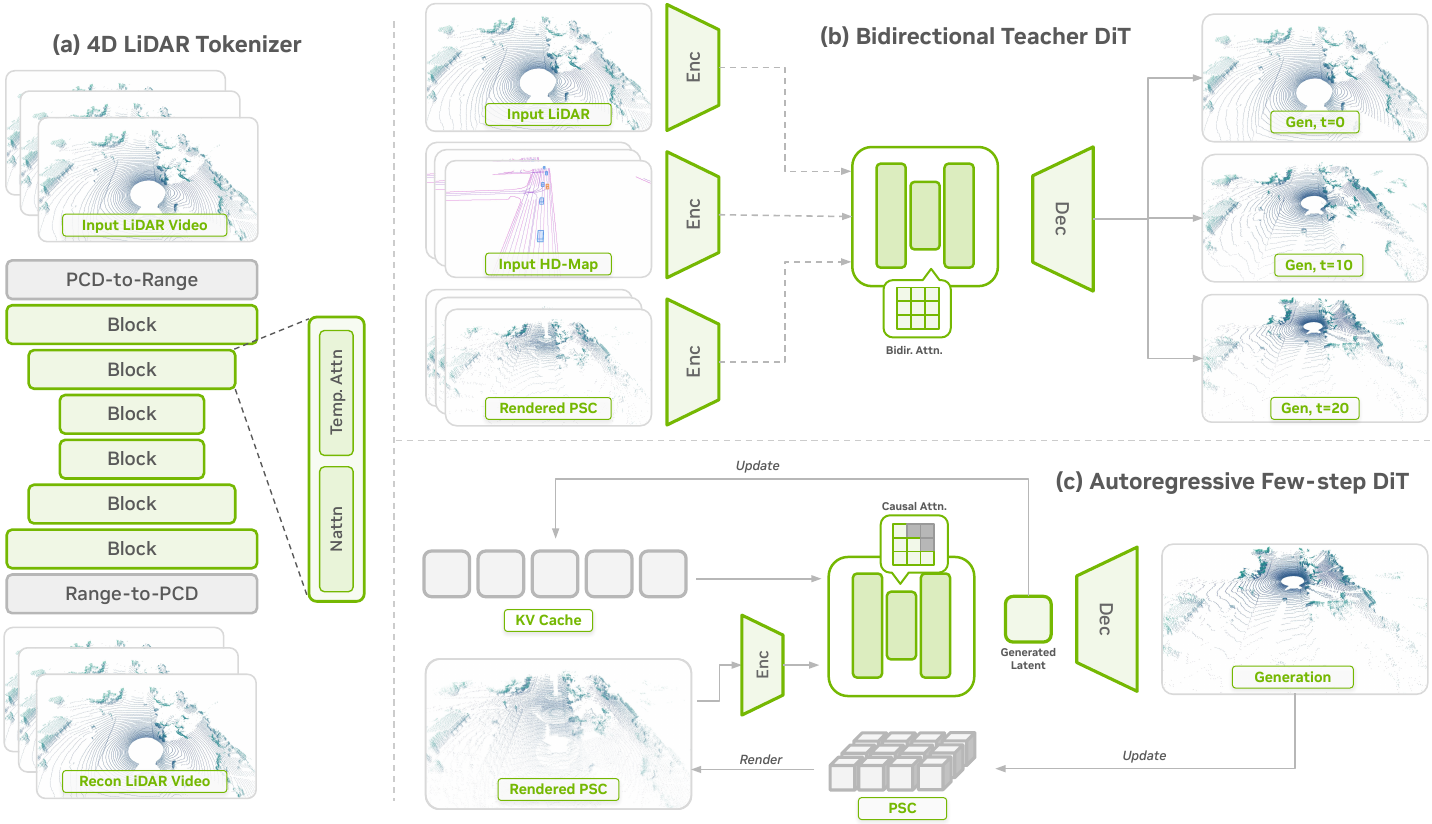}
    \caption{\textbf{Architecture}: \methodname{} is trained in three stages that address representation, task unification, and efficient deployment. (a) A fully attention-based LiDAR tokenizer learns a common latent space while preserving spatial and temporal structure (Sec.~\ref{sec:method-vae}). (b) A bidirectional teacher learns a unified interface for conditioning on a LiDAR prefix, an HD-Map, a persistent spatial cache, and optionally, text (Sec.~\ref{sec:method-bidir}). (c) Causal distillation and a rolling persistent spatial cache enable real-time, long-horizon inference with the same conditioning interface (Sec.~\ref{sec:method-causal}).}
    \label{fig:method}
\end{figure}

Fig.~\ref{fig:method} provides an overview of \methodname{}. Unifying LiDAR sensor simulation requires three capabilities: a compact representation that preserves sparse metric geometry across sensors, a generative model that can operate under different amounts of observed scene evidence, and an efficient recurrent model that remains coherent over long rollouts. We address these requirements in three stages. We first learn a LiDAR-native tokenizer that preserves spatial and temporal structure (Sec.~\ref{sec:method-vae}). We then train a bidirectional teacher with a unified conditioning interface (Sec.~\ref{sec:method-bidir}) and distill it into a few-step causal student with complementary temporal and spatial memory (Sec.~\ref{sec:method-causal}). At inference, changing only the available conditions allows the same causal checkpoint to perform LiDAR generation, point-cloud forecasting, and generalizable LiDAR resimulation (Sec.~\ref{sec:method-inference}).

\subsection{LiDAR-Native Tokenizer}
\label{sec:method-vae}
The unified model first requires a latent representation that is efficient for generation without discarding the sparse geometry or native temporal sampling of LiDAR. It must also accommodate sensors with different scanning patterns within a common interface. We meet these requirements by representing each sweep as a range image and encoding sequences with a fully attention-based tokenizer.

\subsubsection{Range-Image Representation}
We represent each LiDAR sweep as a range image, a parameter-efficient 2.5D representation that is compatible with existing video-generation infrastructure.
A spinning LiDAR measures radial range along a set of calibrated rays arranged into an $H\times W$ grid, where column $i$ represents an azimuth angle $\theta_i$ and row $j$ represents an elevation angle $\phi_j$.
The pixel value $r_{j,i}$ stores the radial range along the corresponding ray, producing a range image $\mathbf R\in\mathbb R^{H\times W}$.
The corresponding 3D point $\mathbf p_{j,i}$ can be computed as
\begin{equation}
  \mathbf p_{j,i}
  =
  \begin{bmatrix}
    x_{j,i} \\
    y_{j,i} \\
    z_{j,i}
  \end{bmatrix}
  =
  r_{j,i}
  \begin{bmatrix}
    \cos\phi_j\cos\theta_i \\
    \cos\phi_j\sin\theta_i \\
    \sin\phi_j
  \end{bmatrix}.
  \label{eq:range_to_3d}
\end{equation}

Conversely, to convert a point cloud $\mathcal P=\{\mathbf p_k\}_{k=1}^{N}$ into a range image, we assign each valid 3D point to its angularly nearest calibrated LiDAR ray. For a point
$\mathbf p_k=[x_k,y_k,z_k]^\top$, its angular coordinates and range are
\begin{align}
    \theta_k=\operatorname{atan2}(y_k,x_k), \qquad
    \phi_k=\arcsin(z_k/r_k), \qquad
    r_k=\lVert\mathbf p_k\rVert_2
    \label{eq:3d_to_range}
\end{align}

If multiple points project to the same ray, we retain the nearest return. We clamp valid ranges to $[r_{\min},r_{\max}]$ and normalize them to $[-1,1]$.
We additionally use a validity mask $\mathbf M\in\{0,1\}^{H\times W}$ to distinguish valid measurements from missing returns. The default range representation concatenates $\mathbf R$ and $\mathbf M$ channel-wise, $\mathbf I=[\mathbf R\mid\mathbf M]$. When intensity is available, we append an intensity image as a third channel after clamping its values to $[0,255]$ and normalizing them to $[0,1]$.

\subsubsection{LiDAR 4D Tokenizer} Stacking $T$ temporally ordered range images produces an input tensor $\mathbf X=\operatorname{Concat}(\mathbf I_1,\ldots,\mathbf I_T)\in\mathbb R^{T\times H\times W\times C_{\mathrm{in}}}$, where $C_{\mathrm{in}}=2$ for range and validity and $C_{\mathrm{in}}=3$ when intensity is included. We train an encoder $\mathcal E$ to map $\mathbf X$ to a latent $\mathbf z=\mathcal E(\mathbf X)\in\mathbb R^{\frac{T}{p_t}\times\frac{H}{p_s}\times\frac{W}{p_s}\times c}$, where $p_s$ and $p_t$ denote the spatial and temporal compression rates, respectively, and $c$ is the latent dimension. A decoder $\mathcal D$ reconstructs the input as $\widehat{\mathbf X}=\mathcal D(\mathbf z)$.

\paragraph{Spatial Neighborhood Attention}
RGB video tokenizers commonly use convolutional backbones~\cite{wan}. In LiDAR range images, however, neighboring angular bins may correspond to surfaces that are meters apart in 3D. Fixed convolutional aggregation can therefore blur depth discontinuities and produce the depth bleeding and wavy boundaries observed in prior work~\cite{liu2026l3dr}.
We instead build our tokenizer from neighborhood attention (NATTEN)~\cite{hassani2023natten}. NATTEN performs content-adaptive aggregation within each local neighborhood while avoiding the quadratic memory cost of global self-attention.

\paragraph{Factorized Spatial and Temporal Attention}
Following latent video diffusion models~\cite{blattmann2023videoldm}, we incorporate temporal context by interleaving each spatial NATTEN block with temporal self-attention. We apply a causal temporal mask so that the representation of each frame depends only on the current and preceding frames, making the tokenizer compatible with causal and variable-length rollouts.

\paragraph{Cross-Sensor Adaptation}
LiDAR streams are typically captured at substantially lower frame rates than RGB video, leaving limited temporal redundancy for compression. We therefore set $p_t=1$ and preserve one latent slice per input sweep. Avoiding temporal downsampling retains frame-level geometry and does not bind the latent representation to a fixed acquisition rate. Together with the fully attention-based backbone, whose content-adaptive aggregation can accommodate different spatial sampling patterns, this design allows a single tokenizer to be jointly adapted across sensors with different beam layouts, angular resolutions, and frame rates.

\paragraph{Optimization}
We train the tokenizer with the following objective:
\begin{align}
  \mathcal L_{\mathrm{tokenizer}} = \lambda_{\mathrm{depth}} \mathcal L_{\mathrm{depth}} + \lambda_{\mathrm{point}} \mathcal L_{\mathrm{point}} + \lambda_{\mathrm{mask}} \mathcal L_{\mathrm{mask}} + \lambda_{\mathrm{KL}} \mathcal L_{\mathrm{KL}},
\end{align}
where $\mathcal L_{\mathrm{depth}}$ denotes an $\ell_1$ loss on valid reconstructed range values, $\mathcal L_{\mathrm{point}}$ measures the $\ell_1$ loss on valid 3D points derived from the reconstructed range image using Eq.~\ref{eq:range_to_3d}, following prior LiDAR generation work~\cite{ran2024lidm}, $\mathcal L_{\mathrm{mask}}$ is the binary cross-entropy loss on the reconstructed validity mask $\widehat{\mathbf M}$, and $\mathcal L_{\mathrm{KL}}$ computes the KL divergence between the latent posterior and a standard normal prior using the reparameterization trick. We set $\lambda_{\mathrm{depth}}=\lambda_{\mathrm{point}}=\lambda_{\mathrm{mask}}=1$ and $\lambda_{\mathrm{KL}}=10^{-6}$. For our final tokenizer, we use $p_s=8$, $p_t=1$, and $c=64$.

\subsection{Bidirectional Teacher Model}
\label{sec:method-bidir}
On the shared tokenizer latent space, we train a single bidirectional diffusion model to learn the conditional distribution of LiDAR sequences. Its conditions encode different forms of scene evidence rather than task identities: a LiDAR prefix provides temporal observations, an HD-Map and actor layout constrain scene structure, a persistent spatial cache anchors previously observed geometry, and text describes environmental appearance. Independently varying these conditions teaches one generative prior to span generation, forecasting, and resimulation before it is transferred to the causal student.

\subsubsection{Base Model}
We initialize the bidirectional teacher from Cosmos-Transfer2.5-2B~\cite{cosmos}. Its diffusion transformer contains 28 blocks with a hidden width of 2048 and 16 attention heads. Each block applies spatiotemporal self-attention, text cross-attention when appropriate, and an MLP with AdaLN-LoRA time conditioning. We retain its 3D rotary positional embeddings and flow-matching formulation~\cite{lipman2022flow,liu2022flow}, while adapting the input and output projections to our LiDAR latent and conditioning channels. Given a clean LiDAR latent $\mathbf z$, we sample noise $\boldsymbol\epsilon\sim\mathcal N(\mathbf 0,\mathbf I_{\mathrm{lat}})$ and a flow time $\tau\in[0,1]$, where $\mathbf I_{\mathrm{lat}}$ is the identity covariance in latent space, and construct
\begin{equation}
  \mathbf z_\tau=(1-\tau)\mathbf z+\tau\boldsymbol\epsilon,
  \qquad
  \mathbf v^*=\frac{\partial\mathbf z_\tau}{\partial\tau}
             =\boldsymbol\epsilon-\mathbf z.
  \label{eq:rectified-flow}
\end{equation}
We train the teacher to predict $\mathbf v^*$ under the unified conditioning interface described below.

\subsubsection{Model Conditions}
The teacher accepts four independently controlled sources of information: a LiDAR prefix, an HD-Map and actor layout, a persistent spatial cache (PSC), and text.

\paragraph{LiDAR Prefix}
The LiDAR prefix contains $K$ observed sweeps preceding the target frames. We convert these sweeps to range images and encode them with the tokenizer. During training and sampling, their clean latents occupy the first $K$ temporal positions and are concatenated temporally with the noisy target latents. A binary mask identifies the observed positions, and the generative loss is applied only to the target positions. This condition enables sequence continuation from one or more LiDAR observations.

\paragraph{HD-Map and Actor Layout}
The layout condition contains lane centerlines, road markings, road boundaries, crosswalks, and 3D actor boxes. Given static map primitives $\mathcal M$, actor boxes $\mathcal B_t$, and target LiDAR-to-world pose $\mathbf T_t$, we transform these elements into the target LiDAR frame and rasterize them into a range-aligned representation:
\begin{equation}
  \mathbf H_t
  =\Pi_{\mathrm{map}}(\mathcal M,\mathcal B_t;\mathbf T_t),
  \label{eq:hdmap-condition}
\end{equation}
where $\Pi_{\mathrm{map}}$ retains the nearest element on each calibrated ray using Eq.~\ref{eq:3d_to_range}. We use the \href{https://github.com/nv-tlabs/Cosmos-Drive-Dreams/blob/main/cosmos-drive-dreams-toolkits/convert_lidar_pointcloud_to_rangemap.py}{preprocessing utility released with Cosmos-Drive-Dreams}. The resulting range image is encoded as $\mathbf h_t=\mathcal E(\mathbf H_t)$ and concatenated channel-wise with the corresponding noisy LiDAR latent.

\paragraph{Persistent Spatial Cache}
The PSC is a world-aligned geometric memory $\mathcal G$ obtained by accumulating point clouds from observations or previous generations. Although its construction differs between the bidirectional and causal models, both use the same conditioning interface. At a target sensor pose $\mathbf T_t$, we project the accumulated geometry into a range image,
\begin{equation}
  \mathbf G_t
  =\Pi_{\mathrm{geometry}}(\mathcal G;\mathbf T_t),
  \label{eq:geom-condition}
\end{equation}
where $\Pi_{\mathrm{geometry}}$ transforms world-frame points into the target LiDAR frame and retains the nearest range on each calibrated ray. We encode this image as $\mathbf g_t=\mathcal E(\mathbf G_t)$ and concatenate it channel-wise with the noisy LiDAR latent.

During teacher training, we construct a fixed PSC from ground-truth sweeps in two modes. The \emph{static-background} mode removes points inside labeled dynamic-object boxes before accumulation. The \emph{frozen-dynamic} mode additionally accumulates each actor in its local coordinate frame and places it according to its box pose at time $t$. The causal model uses the same projection, encoding, and injection procedure, but builds $\mathcal G$ online from generated history as described in Sec.~\ref{sec:method-causal}.

\paragraph{Text}
Following Cosmos-Drive-Dreams~\cite{cosmosdrivedreams}, we use Qwen2.5-VL-7B-Instruct~\cite{qwen2.5} to rewrite scene descriptions into captions with diverse weather and appearance attributes and Cosmos-Reason1 text encoder~\cite{cosmosreason1} to map each caption to token embeddings, which are provided as cross-attention context in every diffusion-transformer block. Text controls environmental attributes, while the spatial conditions anchor scene geometry.

\subsubsection{Condition Sampling}
\begin{figure}[H]
\begin{minipage}{\linewidth}
\paragraph{Text-Controlled Weather Generation}
Finally, we show text-controlled weather generation on the NVIDIA PAI dataset in Fig.~\ref{fig:weather-cond}. We use Qwen2.5-VL-7B-Instruct~\cite{qwen2.5} to produce detailed weather descriptions from front-view camera images and Cosmos-Reason1~\cite{cosmosreason1} to encode the text into embeddings and condition the bidirectional model through cross-attention at every DiT layer. The model reproduces typical rainy-weather LiDAR patterns, including increased ray drop on the road surface.
\par\medskip
\centering
    \includegraphics[width=1\linewidth]{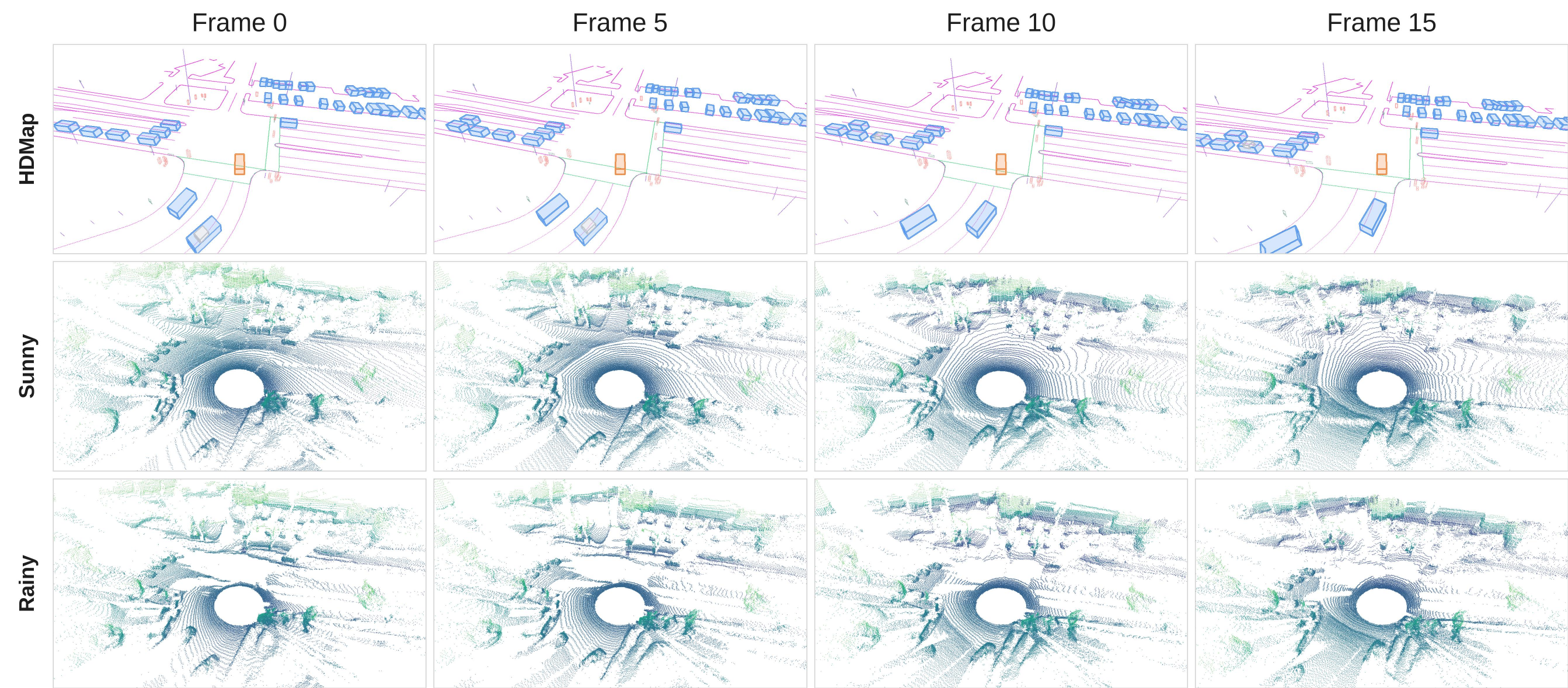}
    \caption{\textbf{Weather-controlled generation with the text-conditioned bidirectional teacher.} We modify the original sunny-scene prompt to request rainy weather.}
    \label{fig:weather-cond}
\end{minipage}
\end{figure}

During the unified-conditioning stage, we sample all conditioning factors independently rather than training task-specific models. With probability 0.5, we sample the \textit{LiDAR prefix} length $K$ uniformly from $\{1,2,4,8,10,12,16\}$; otherwise, $K=0$. We drop the \textit{HD-Map} condition with probability 0.5 using a null token. For the \textit{PSC}, we sample null geometry, static-background geometry, and frozen-dynamic geometry with probabilities $0.1$, $0.45$, and $0.45$, respectively. In the frozen-dynamic mode, the number of source sweeps $N_{\mathcal G}$ is sampled uniformly from $\{1,2,4,8,10,12,16,20\}$. For the text-conditioned results in Fig.~\ref{fig:weather-cond}, we use 20\% classifier-free dropout of the text embeddings. This factorized sampling exposes the model to diverse combinations of scene evidence without task-specific tokens or objectives, producing the shared conditional prior used for causal distillation.

\begin{figure}[!t]
    \centering
    \includegraphics[width=1\linewidth]{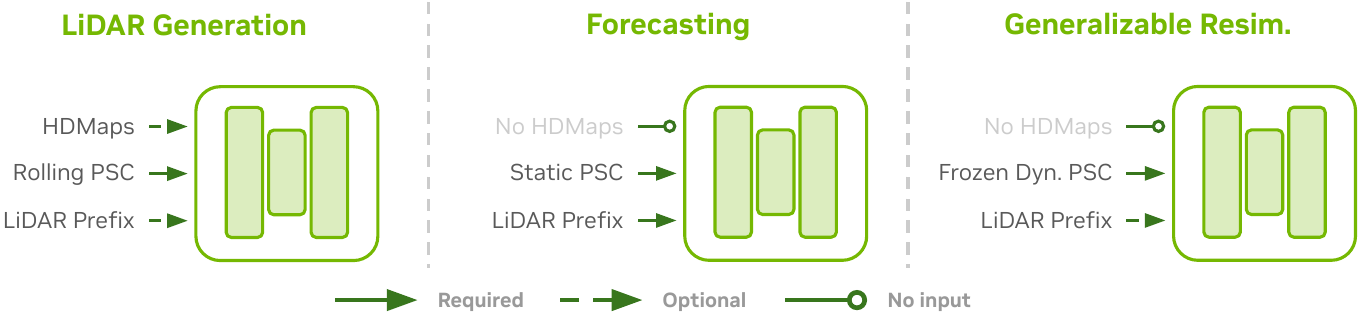}
    \caption{\textbf{Unified inference}: the same causal checkpoint performs LiDAR generation, point-cloud forecasting, and generalizable LiDAR resimulation through different condition combinations. All tasks additionally receive the requested target sensor poses; gray labels denote omitted conditions.}
    \label{fig:conditioning}
\end{figure}

\subsection{Causal Distillation}
\label{sec:method-causal}
The bidirectional teacher provides a flexible, high-quality generative prior, but its full-sequence attention and iterative denoising are unsuitable for interactive simulation. We therefore distill it into a few-step, block-autoregressive causal student and equip the student with complementary temporal and spatial memories for long-horizon rollout. The student retains the conditioning interface from Sec.~\ref{sec:method-bidir}, allowing efficiency to improve without changing how the simulation tasks are specified.

\subsubsection{Causal Initialization}
We initialize the causal student from the trained bidirectional teacher and replace full temporal attention with a block-causal pattern. To support generation from a single frame, we use a singleton attention sink~\cite{xiao2024streamingllm} followed by four-frame blocks. During causal initialization, clean ground-truth latents from preceding blocks are stored in a K/V cache, while each active block is independently perturbed and denoised. This stage adapts the teacher weights to causal attention before self-forcing distillation.

\subsubsection{Self-Forcing Distillation}
We next adopt self-forcing distribution matching distillation (SF-DMD)~\cite{huang2025selfforcing,yin2024dmd}.
Starting from the causal initialization, the causal student sequentially samples each four-frame block with four shifted rectified-flow SDE steps, commits the generated clean latents to its K/V cache, and updates the recurrent conditioning state before sampling the next block.
Training therefore uses the same self-generated temporal and geometric histories encountered at inference.

Rather than regressing the rollout to a paired ground-truth sequence, SF-DMD
matches its distribution to the real LiDAR latent distribution. Let
$\widetilde{\mathbf z}$ denote a self-rolled-out latent sequence. After perturbing
it to $\widetilde{\mathbf z}_\tau$, a frozen real-data denoiser and a learned
generated-data denoiser produce clean-latent estimates
$\widehat{\mathbf z}_0^{\mathrm{real}}$ and
$\widehat{\mathbf z}_0^{\mathrm{gen}}$, respectively. For a replayed denoising
step, define the adaptive scale
$a=\operatorname{mean}|\widetilde{\mathbf z}
-\widehat{\mathbf z}_0^{\mathrm{real}}|+\varepsilon$. The student's clean prediction
$\widehat{\mathbf z}_0^{\mathrm{stu}}$ is optimized with
\begin{equation}
  \begin{aligned}
    \mathcal L_{\mathrm{SF\text{-}DMD}}
    =\mathbb E\!\left[\frac{1}{2}\left\|
      \widehat{\mathbf z}_0^{\mathrm{stu}}
      -\operatorname{sg}\!\left[
      \widetilde{\mathbf z}
      -\frac{\widehat{\mathbf z}_0^{\mathrm{gen}}
      -\widehat{\mathbf z}_0^{\mathrm{real}}}{a}
      \right]
    \right\|_2^2\right].
  \end{aligned}
  \label{eq:sfdmd-loss}
\end{equation}
where $\operatorname{sg}$ denotes stop-gradient and $a$ adaptively normalizes
the update. The real-data denoiser remains frozen, while the generated-data
denoiser is initialized from it and trained with the rectified-flow objective on
self-rollouts. The conditioning factors remain independently randomized, so
the distilled student retains the unified task interface learned by the
teacher.

\subsubsection{Rolling Persistent Spatial Cache}
A finite temporal K/V cache retains only recent latent context; increasing its length incurs growing memory and computational cost~\cite{gu2025far,xi2026quantvideogen,samuel2026fastar}. Once an observation leaves this window, the model may lose previously generated scene structure~\cite{guo2026memorize,wu2026worldtrace}. We therefore maintain a sparse, world-aligned voxel map $\mathcal V$ as a rolling PSC. It uses the same PSC conditioning interface defined in Sec.~\ref{sec:method-bidir}, but constructs the underlying geometry online from generated history.

After generating block $b$, we decode each latent $\widetilde{\mathbf z}_t$ into a point cloud $\widetilde{\mathcal X}_t$, register it to the common world frame using the corresponding sensor pose, and update the voxel map:
\begin{equation}
  \mathcal V_b
  =\mathcal U(\mathcal V_{b-1},\mathcal Y_b),\qquad
  \mathcal Y_b
    =\bigcup_{t\in\mathcal T_b}
      \mathbf T_t\widetilde{\mathcal X}_t,
  \label{eq:spatial-cache-register}
\end{equation}
where $\mathcal T_b$ contains the frame indices in block $b$ and $\mathcal U$ is the cache update operator. Each occupied voxel $v$ stores its most recently registered point $\mathbf p(v)$, an observation count $n^{\mathrm{obs}}(v)$, and a removal count $n^{\mathrm{rem}}(v)$. Each source frame contributes at most one observation to a voxel. When a voxel is observed, we update its point, increment its observation count, and reset its removal count. When valid free-space evidence contradicts a previously occupied voxel, we increment its removal count. Requiring repeated observations prevents transient generation errors from entering persistent memory, while requiring repeated free-space contradictions prevents a single inconsistent sweep from deleting valid structure.

Before generating block $b+1$, we retain only confirmed, non-removed voxels and render them at each requested target pose:
\begin{equation}
  \begin{aligned}
  \mathcal P(\mathcal V_b)
    &=\left\{\mathbf p(v)\mid
      n^{\mathrm{obs}}(v)\geq N_{\min},\;
      n^{\mathrm{rem}}(v)<N_{\mathrm{rem}}\right\},\\
  \mathbf g_t^{\mathrm{cache}}
    &=\mathcal E\!\left(
      \Pi_{\mathrm{geometry}}(\mathcal P(\mathcal V_b);\mathbf T_t)
    \right),\qquad t\in\mathcal T_{b+1}.
  \end{aligned}
  \label{eq:spatial-cache-query}
\end{equation}
Thus each block queries geometry produced only by earlier blocks. The new outputs are registered with Eq.~\ref{eq:spatial-cache-register} only after the block has been generated, preventing self-conditioning and future leakage. During SF-DMD and inference, this rolling cache replaces the fixed ground-truth PSC used during bidirectional and causal-initialization training without changing the conditioning channel or model architecture.

\paragraph{Cache Implementation}
We use $0.15\,\mathrm{m}$ cubic voxels, each storing its latest point, observation count, and removal count. Each generated frame increments the observation count at most once per voxel. In case of repeated observation within a voxel, we retain the latest point and reset the removal count. To detect stale geometry, we project eligible cached points into the last sweep's range image. If the surrounding $3\times3$ pixels contain valid returns and are farther from the cached point by more than $0.75\,\mathrm{m}$, we increment its voxel's removal count.
For conditioning, only voxels with at least $N_{\min}=3$ observations contribute to the rendered PSC. Voxels reaching $N_{\mathrm{rem}}=2$ contradictions are deleted together with their counts. If those voxels are observed again, they are registered anew and must satisfy the observation threshold again. Cache updates occur only after block generation, preventing a block from conditioning on its own outputs.

\subsection{Model Inference}
\label{sec:method-inference}

At inference, the conditions organize sensor simulation along a spectrum from new-world synthesis to reconstruction-grounded rendering, as summarized in Fig.~\ref{fig:conditioning}. Every mode receives the requested target sensor poses, and an HD-Map may additionally constrain the scene layout. Increasing the observed LiDAR evidence through a prefix or PSC moves the same causal checkpoint from generating a new world to forecasting an observed sequence and finally to resimulating a particular captured scene.

\paragraph{LiDAR Generation}
The model may receive an HD-Map sequence and may start from a LiDAR prefix; it then builds its rolling PSC from generated sweeps. For interactive action-conditioned generation, user-provided ego actions determine the future sensor poses.

\paragraph{Point-Cloud Forecasting}
Observed LiDAR history initializes the LiDAR prefix, temporal K/V cache, and PSC. Without an HD-Map condition, the model predicts future sweeps at the provided ego poses while updating both caches from generated history.

\paragraph{Generalizable LiDAR Resimulation}
Sparse observed sweeps are accumulated into a fixed frozen-dynamic PSC and rendered at novel target sensor poses. An observed LiDAR prefix can optionally provide additional temporal context; no HD-Map condition or per-scene optimization is required.

\section{Experiments}
\label{sec:experiments}

We organize the evaluation around four questions. First, does the LiDAR-native tokenizer preserve metric geometry while producing latents that are effective for generative modeling? Second, can one bidirectional teacher express diverse behaviors through the shared conditioning interface? Third, does the distilled causal checkpoint retain this interface while operating in real time across autoregressive LiDAR generation, point-cloud forecasting, and generalizable LiDAR resimulation? Finally, which architectural and memory components account for its performance? We primarily evaluate on the large-scale Waymo dataset~\cite{sun2020waymo}, then study intensity modeling and generalization across sensors and datasets. Unless stated otherwise, \methodname{} denotes the causal student configured according to Sec.~\ref{sec:method-inference}; results from the bidirectional teacher are identified explicitly. The cross-sensor, intensity, and text-controlled weather extensions use fine-tuned bidirectional teachers.

\subsection{Implementation Details}
\label{sec:implementation-details}

\paragraph{Tokenizer}
Our tokenizer is a symmetric Transformer with channel widths $[128,256,512]$, spatial depths $[4,2,4]$, and $[4,16,16]$ attention heads, corresponding to ten spatial Transformer blocks in each of the encoder and decoder. The spatial NATTEN~\citep{hassani2023natten} blocks use a $5\times35$ neighborhood window with no dilation and circular padding, following \cite{nakashima2024r2flow}, to handle azimuthal discontinuities. Factorized temporal-attention blocks are applied at the two lower-resolution stages, while no temporal downsampling is performed. The tokenizer uses a $2\times2$ input patch embedding, an MLP expansion ratio of $3$, two latent-mapping layers, and a $64$-channel latent space. It compresses each sweep spatially by $8\times8$, mapping an input of size $2\times64\times1024$ to a latent of size $64\times8\times128$ while preserving temporal resolution. The decoder predicts both the range image and its validity mask. We train the tokenizer for $120\,000$ iterations. The total parameter count for the tokenizer is 64.5M.  We set $\lambda_{\mathrm{depth}}=\lambda_{\mathrm{point}}=\lambda_{\mathrm{mask}}=1$ and $\lambda_{\mathrm{KL}}=10^{-6}$. For our final tokenizer, we use $p_s=8$, $p_t=1$, and $c=64$.

\paragraph{Bidirectional and Causal Model Training}
Our DiT is initialized from Cosmos-Transfer2.5-2B~\citep{cosmos} and subsequently adapted to the LiDAR latent space. It contains 28 blocks with a hidden width of 2048 and 16 attention heads. Each block applies spatiotemporal self-attention, text cross-attention when appropriate, and an MLP with AdaLN-LoRA time conditioning. We retain the 3D rotary positional embeddings and flow-matching formulation~\citep{lipman2022flow,liu2022flow}.  
We train the world model in three stages. \textit{First}, we train the bidirectional flow-matching teacher for $270\,000$ iterations: $150\,000$ iterations of HD-Map-conditioned generation on nine-frame clips, $100\,000$ iterations on 21-frame clips, and $20\,000$ iterations of unified conditioning on 21-frame clips. Empirically, this curriculum produced better results than directly training with joint conditioning. \textit{Second}, we initialize the causal student from the teacher and train it with teacher forcing on 21-frame clips for $10\,000$ iterations, using one prefix frame followed by causal blocks of four frames. \textit{Finally}, we perform self-forcing distribution matching on 21-frame clips for $2\,000$ iterations. We update the fake-score network every cycle and the student every five cycles. The final causal student generates each four-frame block using four steps and maintains a rolling nine-frame K/V cache including one persistent sink frame. The four-step SDE noise schedule is $(1,15/16,5/6,5/8,0)$, with zero denoting the clean endpoint. Table~\ref{tab:training-config} lists the training configurations.

\begin{table}[!htbp]
  \centering
  \caption{\textbf{Training configurations for our three-stage pipeline.}}
  \label{tab:training-config}
  \small
  \renewcommand{\arraystretch}{1.15}
  \setlength{\tabcolsep}{12pt}
  \resizebox{\textwidth}{!}{
  \begin{tabular}{lccc}
    \toprule
    \textbf{Configuration}
    & \makecell{\textbf{Bidirectional} \textbf{training}}
    & \makecell{\textbf{Causal} \textbf{initialization}}
    & \makecell{\textbf{Self forcing} \textbf{distillation}} \\
    \midrule
    Global batch size & 32 & 32 & 32 \\
    Training clip length & 9 $\rightarrow$ 21 frames & 21 frames & 21 frames \\
    \addlinespace
    Model optimizer
    & \makecell{AdamW\\
        $\epsilon=10^{-8}$ \\
        $\mathrm{lr}=10^{-5}$\\
        $\beta=(0.9,0.999)$\\
        $\mathrm{wd}=0.01$}
    & \makecell{AdamW\\
        $\epsilon=10^{-8}$ \\
        $\mathrm{lr}=10^{-5}$\\
        $\beta=(0.9,0.999)$\\
        $\mathrm{wd}=0.01$}
    & \makecell{AdamW\\
        $\epsilon=10^{-8}$ \\
        $\mathrm{lr}=2\times10^{-6}$\\
        $\beta=(0.9,0.999)$\\
        $\mathrm{wd}=0.01$} \\
    \addlinespace
    Fake-score optimizer & -- & --
    & \makecell{AdamW\\
        $\epsilon=10^{-8}$ \\
        $\mathrm{lr}=4\times10^{-7}$\\
        $\beta=(0.9,0.999)$\\
        $\mathrm{wd}=0.01$} \\
    \addlinespace
    RF timestep shift & 5 & 5 & 5 \\
    Distillation teacher CFG & -- & -- & 3.0 \\
    Student sampling & -- & -- & Four-step SDE, $\eta=1$ \\
    Training iterations & 270,000 & 10,000 & 2,000 \\
    \bottomrule
  \end{tabular}
  }
\end{table}

\paragraph{Datasets}
We use Waymo Open Dataset~\cite{sun2020waymo} for model training and evaluation. Its training split has 798 sequences and validation split has 202 sequences. Each sequence contains approximately 20 seconds of LiDAR captured at 10\,Hz, giving roughly 160K training sweeps, 4.4 hours of sensor data, and 142.8K possible overlapping 21-frame training clips. Both the tokenizer and all three stages of world-model training use this split. For the cross-sensor experiments in Sec.~\ref{sec:cross-dataset}, we additionally fine-tune the tokenizer and bidirectional teacher on NVIDIA PAI~\cite{nvidiapai}, Waymo, and nuScenes~\cite{nuscenes}. This mixture contains approximately 640K unique 21-frame clips from 111,988 NVIDIA PAI sequences, 798 Waymo sequences, and 722 nuScenes sequences. To prevent the substantially larger NVIDIA PAI corpus from dominating training, we sample the three datasets with equal probability.

\subsection{Tokenizer Performance}
\label{sec:vae-perf}
The tokenizer is the common representation for both generative simulation and reconstruction-grounded resimulation. We therefore evaluate not only its ability to reconstruct sparse metric geometry, but also how readily its latents can be modeled by a fixed diffusion transformer and how efficiently they can be encoded and decoded.

\paragraph{Baselines} We study two compression regimes. At a matched $256\times$ spatiotemporal compression rate, we compare against the Cosmos CV~\cite{cosmos-tokenizer} and Wan2.1~\cite{wan} video tokenizers, both of which compress time by $4\times$. At lower compression rates of $32\times$ to $64\times$, we compare against the LiDM tokenizer~\cite{ran2024lidm} and the tokenizer used by Cosmos LiDARGen~\cite{cosmosdrivedreams}. We retrain all baselines from scratch using the same range-image representation. To isolate the contribution of the latent representation to downstream generation, we pair every tokenizer with the same DiT architecture, training recipe, and evaluation protocol. The tokenizer reconstruction comparison and the corresponding DiT training and generation evaluation use nine-frame sequences.

\paragraph{Metrics} We evaluate reconstruction using Chamfer distance (CD), range $L_1$, and absolute relative error (AbsRel). We assess distributional generation quality using BEV-based Maximum Mean Discrepancy (MMD) and Jensen--Shannon divergence (JSD), following \citet{zyrianov2022lidargen}, and feature-space quality using Fr\'echet Utonia Distance (FUD), computed with the universal point-cloud encoder Utonia~\cite{zhang2026utonia}. We measure temporal consistency using Temporal Transformation Consistency Error (TTCE) and Chamfer Temporal Consistency (CTC) from \citet{liu2025lidarcrafter}, together with ICP-Avg. from \citet{zyrianov2024lidardm}.

\begin{table}[!htbp]
\centering
\caption{\textbf{Tokenizer comparison using the same DiT.} ST denotes
spatiotemporal compression. Reconstruction and generation use nine-frame sequences. MMD values are multiplied by 10$^5$.}
\label{tab:vae-comparison}
\scriptsize
\setlength{\tabcolsep}{2.2pt}
\resizebox{\textwidth}{!}{%
\begin{tabular}{l c ccc ccc cc cc cc}
\toprule
\multirow{2}{*}{Method} & \multirow[c]{2}{*}{\makecell[c]{ST Comp.\\rate}} & \multicolumn{3}{c}{Reconstruction}
& \multicolumn{3}{c}{Generation} & \multicolumn{2}{c}{TTCE $\downarrow$}
& \multicolumn{2}{c}{CTC $\downarrow$}
& \multirow[c]{2}{*}{\makecell[c]{ICP Avg.~$\downarrow$}}
& \multirow[c]{2}{*}{\makecell[c]{TFLOPs\\/frame $\downarrow$}} \\
\cmidrule(lr){3-5}\cmidrule(lr){6-8}\cmidrule(lr){9-10}\cmidrule(lr){11-12}
&  & CD $\downarrow$ & $L_1$ $\downarrow$ & AbsRel $\downarrow$
& JSD $\downarrow$ & MMD $\downarrow$ & FUD $\downarrow$
& 3 & 4 & 3 & 4 & & \\
\midrule
\multicolumn{14}{l}{\textit{ST compression rate: 256x}} \\
Cosmos CV~\cite{cosmos-tokenizer} & 8x8x4
& 0.439 & 0.790 & 0.031
& \underline{0.061} & \underline{6.41} & 46.26
& \underline{1.773} & \underline{2.561}
& \underline{0.322} & \textbf{0.357}
& 0.089 & \textbf{0.058} \\
Wan2.1~\cite{wan} & 8x8x4
& \underline{0.297} & \underline{0.395} & \underline{0.017}
& \textbf{0.057} & \textbf{5.72} & \underline{37.52}
& 1.983 & 2.832
& 0.337 & 0.381
& \underline{0.077} & 0.911 \\
\textbf{Ours} & 16x16x1
& \textbf{0.175} & \textbf{0.143} & \textbf{0.008}
& \underline{0.061} & 6.87 & \textbf{32.84}
& \textbf{1.729} & \textbf{2.492}
& \textbf{0.311} & \textbf{0.357}
& \textbf{0.072} & \underline{0.400} \\
\midrule
\multicolumn{14}{l}{\textit{ST compression rate: 32x / 64x}} \\
LiDM~\cite{ran2024lidm} & 4x8x1
& \underline{0.297} & \underline{0.371} & \underline{0.016}
& \textbf{0.057} & \textbf{5.48} & \underline{36.22}
& \underline{1.696} & \underline{2.477}
& 0.315 & 0.352
& 0.088 & \textbf{0.212} \\
Cosmos LiDARGen~\cite{cosmosdrivedreams} & 8x8x1
& 0.384 & 0.609 & 0.025
& \underline{0.062} & \underline{7.99} & 85.37
& 1.738 & 2.523
& \underline{0.304} & \textbf{0.338}
& \underline{0.085} & 0.302 \\
\textbf{Ours} & 8x8x1
& \textbf{0.115} & \textbf{0.072} & \textbf{0.005}
& 0.065 & 9.14 & \textbf{18.17}
& \textbf{1.681} & \textbf{2.463}
& \textbf{0.294} & \underline{0.339}
& \textbf{0.059} & \underline{0.250} \\
\bottomrule
\end{tabular}%
}
\end{table}

\paragraph{Results} Tab.~\ref{tab:vae-comparison} shows that our LiDAR-native tokenizer preserves geometry substantially more faithfully in both compression regimes. At $256\times$ compression, our temporally uncompressed design reduces CD by $41\%$ and AbsRel by $53\%$ relative to the strongest competing reconstruction baseline, while also obtaining the best FUD and nearly all temporal metrics. At lower compression, our $8{\times}8{\times}1$ tokenizer reduces CD by $61\%$ and FUD by $50\%$ relative to the best competing result, despite LiDM using a lower $32\times$ compression rate. Preserving native temporal sampling with a LiDAR-specific architecture therefore produces a shared latent space that is both geometrically faithful and substantially easier to model generatively.

\subsection{Bidirectional Teacher Model}
\label{sec:bidir-teacher-results}

Before evaluating real-time autoregressive inference, we test whether the bidirectional teacher has learned a useful shared conditional prior. Because it attends to the full sequence during denoising, it isolates the representational capacity and conditioning flexibility of the base model from errors introduced by causal rollout. The unconditional, HD-Map-conditioned, and first-frame-conditioned results use the teacher checkpoint from Sec.~\ref{sec:method-bidir}; the weather-control example uses its text-conditioned NVIDIA PAI extension.

\begin{figure}[H]
\begin{minipage}{\linewidth}
\paragraph{HD-Map-Conditioned Generation}
To assess condition alignment, we present HD-Map-conditioned generation results from our teacher in Fig.~\ref{fig:bidir-hdmap}. The model generates 21 frames conditioned on corresponding 21-frame HD-Map sequences. It follows the conditioning well across highway (Sample 0), pedestrian-rich (Sample 1), and urban (Sample 2) scenarios.
\par\medskip
\centering
    \includegraphics[width=1\linewidth]{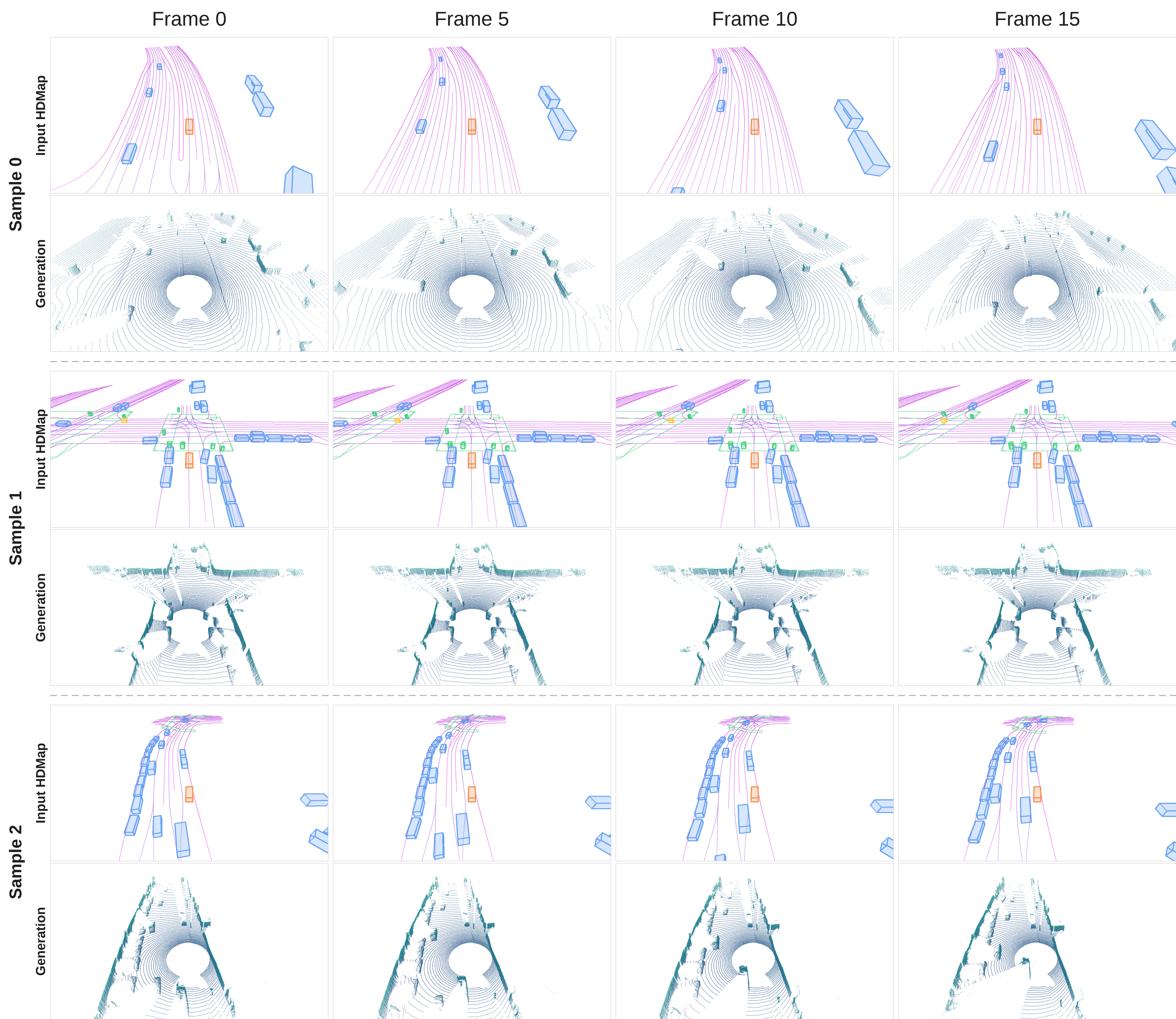}
    \caption{\textbf{HD-Map-conditioned generation with the bidirectional model.} Sample 0 depicts a high-speed highway, Sample 1 a dense intersection with pedestrians, and Sample 2 a moving truck approaching from behind. Each sequence is conditioned only on the HD-Map.}
    \label{fig:bidir-hdmap}
\end{minipage}
\end{figure}

\begin{figure}[H]
\begin{minipage}{\linewidth}
\paragraph{Unconditional Generation}
We present unconditional generation results from our bidirectional teacher in Fig.~\ref{fig:bidir-uncond}. The model generates realistic and diverse 4D LiDAR sequences from noise, ranging from high-speed highway scenarios (Sample 0) to dense urban traffic (Sample 1).
\par\medskip
\centering
    \includegraphics[width=1\linewidth]{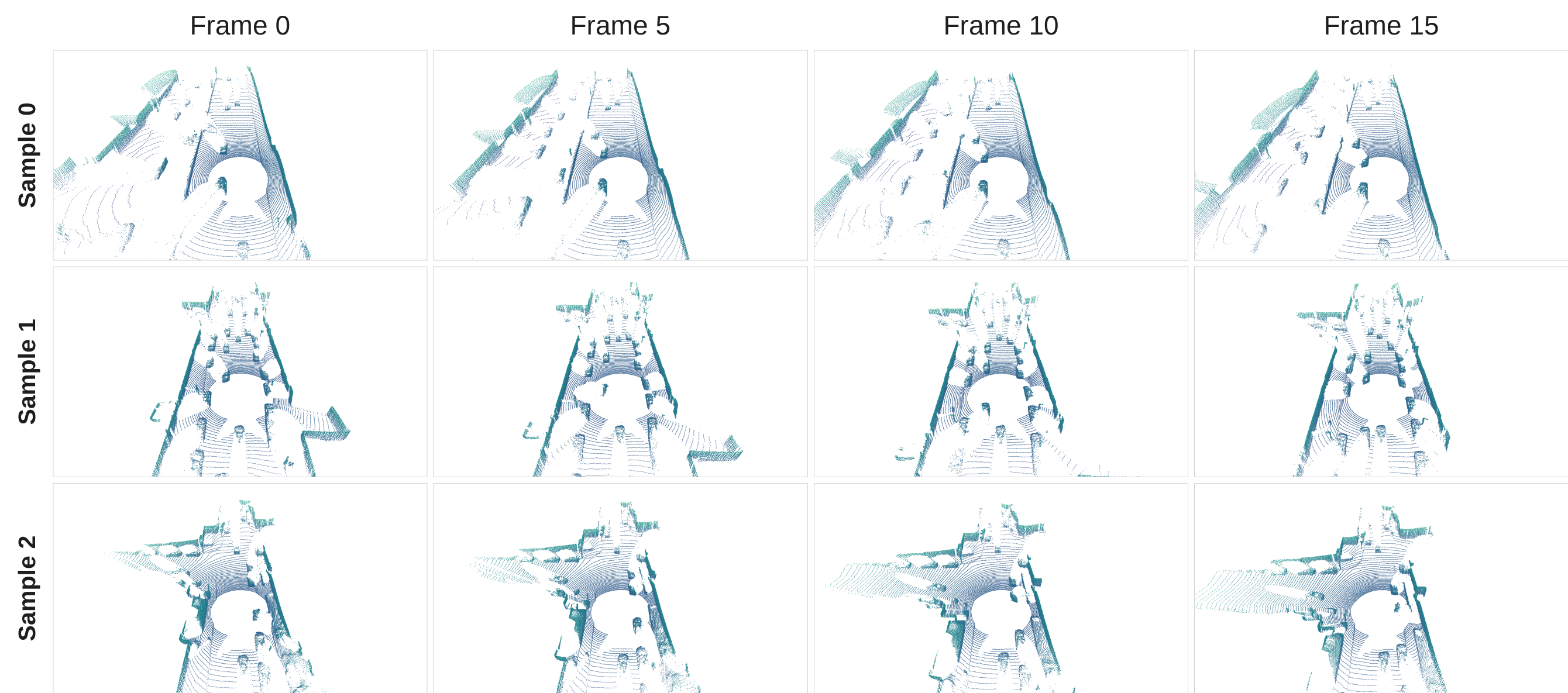}
    \caption{\textbf{Unconditional generation with the bidirectional model.} \methodname{} produces high-quality and diverse unconditional samples. Sample 0 depicts a highway scene, Sample 1 dense traffic, and Sample 2 an intersection. All conditioning inputs are set to null.}
    \label{fig:bidir-uncond}
\end{minipage}
\end{figure}

\begin{figure}[H]
\begin{minipage}{\linewidth}
\paragraph{First-Frame-Conditioned Generation}
We further show first-frame-conditioned generation in Fig.~\ref{fig:bidir-firstframe}. Given one observed LiDAR frame, the model generates a continuation of the sequence. The resulting sequences contain plausible future motion in which agents follow the traffic structure in the observed scene.
\par\medskip
\centering
    \includegraphics[width=1\linewidth]{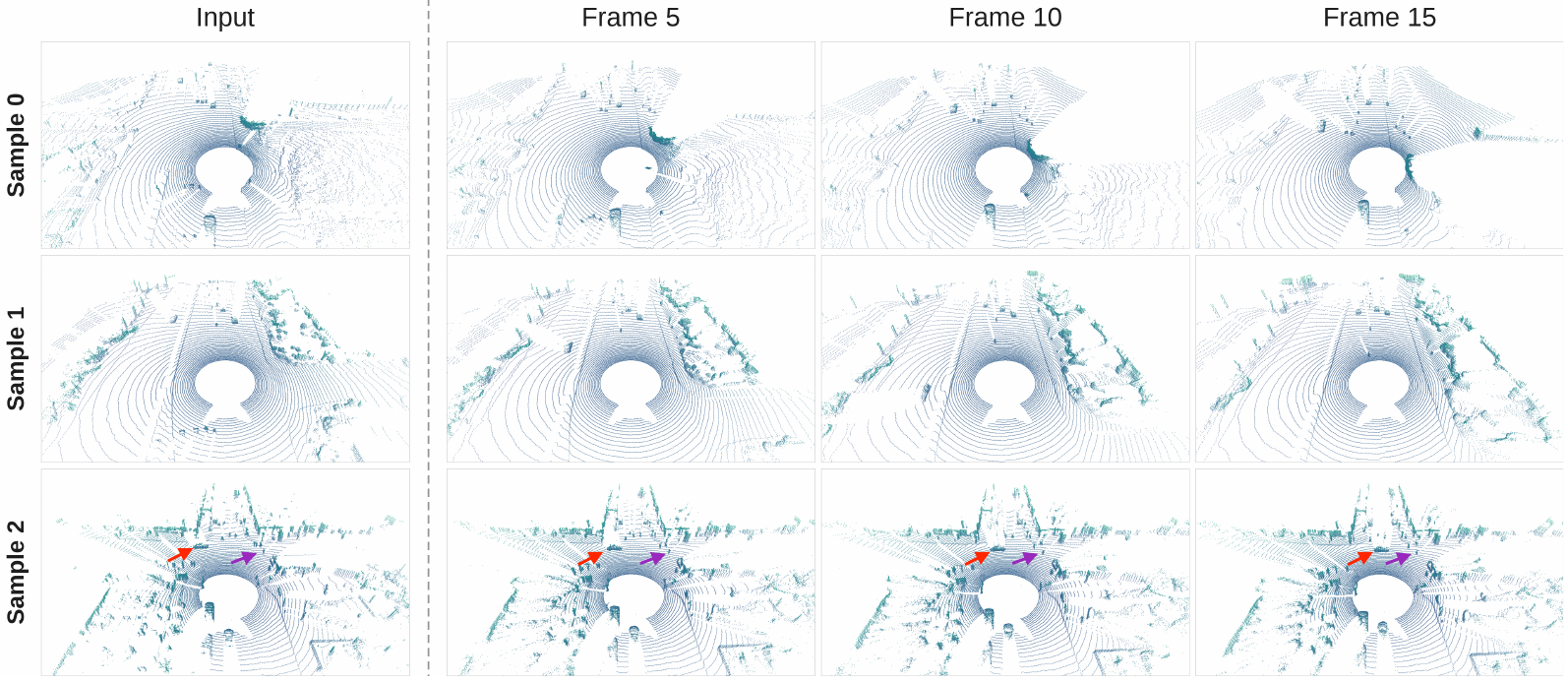}
    \caption{\textbf{First-frame-conditioned generation with the bidirectional model.} The red arrow marks a vehicle crossing the intersection, and the purple arrow marks a pedestrian in the crosswalk. Sample 0 depicts a high-speed highway, Sample 1 urban traffic, and Sample 2 a crowded intersection with a pedestrian crossing the road.}
    \label{fig:bidir-firstframe}
\end{minipage}
\end{figure}

Together, these results show that one bidirectional prior can synthesize diverse scenes while responding to temporal, geometric, and semantic evidence through a common interface. We next test whether causal distillation preserves this flexibility during real-time, long-horizon rollout.

\subsection{Autoregressive LiDAR Generation}
\label{sec:causal-generation}
We now evaluate the same causal checkpoint across the three inference regimes introduced in Sec.~\ref{sec:method-inference}, beginning with autoregressive generation. This is the least reconstruction-constrained regime: the model must synthesize a coherent world over long horizons from scene-level conditions and, optionally, an initial LiDAR observation.

\begin{figure}[!htbp]
    \centering
    \includegraphics[width=1\linewidth]{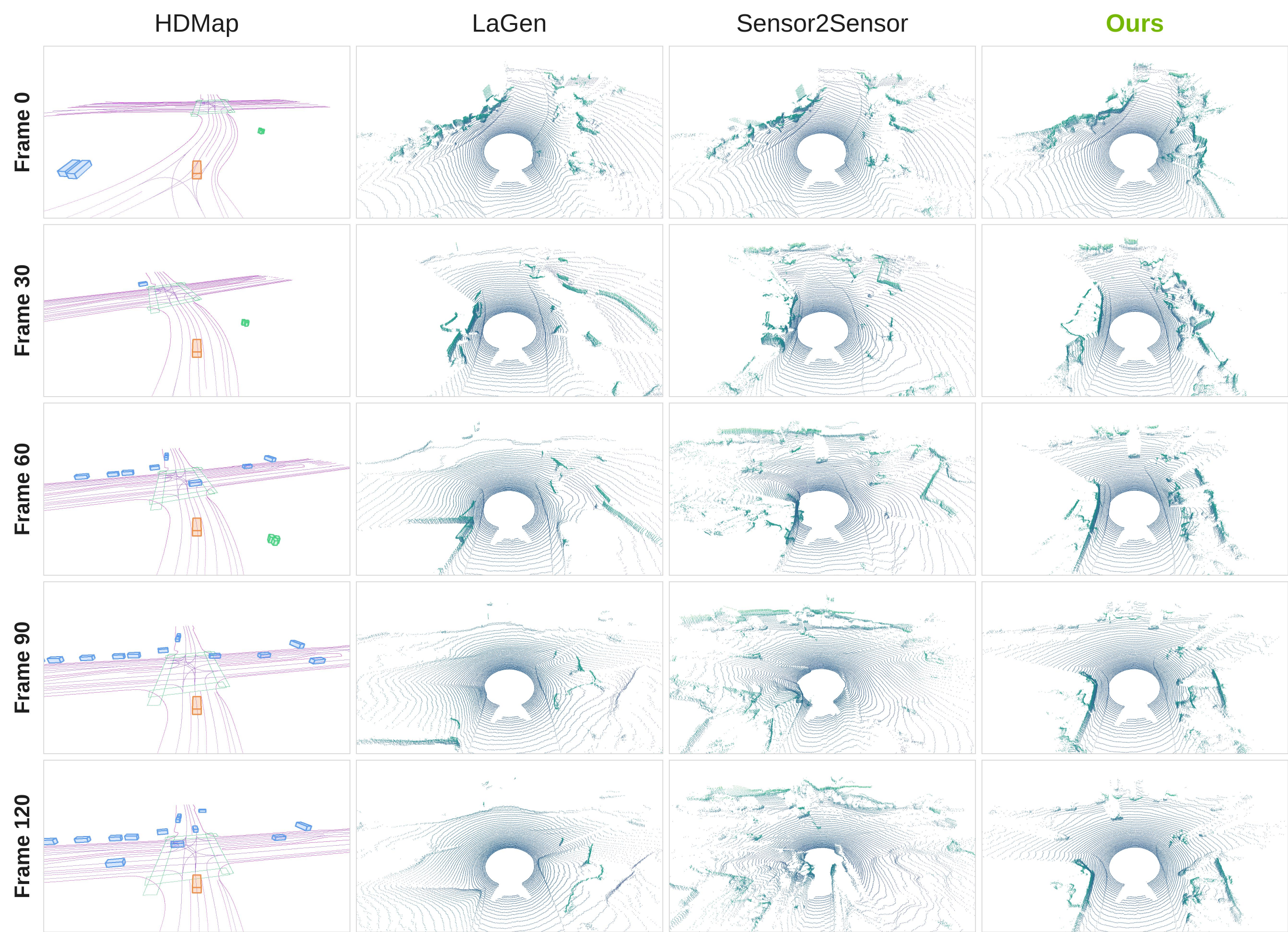}
    \caption{\textbf{Autoregressive LiDAR generation}: \methodname{} generates realistic, temporally consistent LiDAR sequences over long horizons, whereas the baselines collapse after 100 frames.
    }
    \label{fig:ar_gen_cond}
\end{figure}

\paragraph{Settings} We evaluate the causal student in two autoregressive LiDAR generation settings: HD-Map-conditioned generation and interactive action-conditioned generation. In the first setting, the model produces 129-frame rollouts at 10\,Hz from the corresponding HD-Map sequences: one generated singleton frame followed by 128 generated frames in 32 four-frame blocks. In the second setting, the first ground-truth frame supplies the singleton and ego actions specify the future sensor trajectory as described in Sec.~\ref{sec:method-inference}; the model generates the following 128 frames, giving 129 frames including the observation.

\paragraph{Metrics} We use the same generation and temporal-consistency metrics defined in Sec.~\ref{sec:vae-perf}. To assess long-horizon stability (Fig.~\ref{fig:causal-degradation}), we compute FUD over non-overlapping eight-frame windows. This windowed FUD balances reliable distributional estimates with sensitivity to rollout degradation.

\paragraph{Baselines} We compare \methodname{} with LiDARCrafter~\cite{liu2025lidarcrafter}, LaGen~\cite{liu2025lagen}, and Sensor2Sensor~\cite{wang2026sensor2sensor}. For LiDARCrafter, we use the released codebase and train the model from scratch under our protocol. Because code was not publicly available for LaGen and Sensor2Sensor when these experiments were conducted, we reimplement both methods using our tokenizer, a matched DiT backbone, and the same training protocol. We adapt the methods to HD-Map and action conditioning where necessary. These reimplemented results are marked with $^*$ in Tab.~\ref{tab:causal-generation}.

\begin{table}[!htbp]
    \centering
    \caption{\textbf{Autoregressive LiDAR generation.} We evaluate the causal \methodname{} checkpoint. MMD values are multiplied by 10$^5$. $^*$ indicates a reimplementation using our tokenizer and a matched DiT.}
    \label{tab:causal-generation}
    \scriptsize
    \setlength{\tabcolsep}{3.2pt}
    \resizebox{\textwidth}{!}{%
    \begin{tabular}{l c ccc cc cccc cc}
    \toprule
    \multirow{2}{*}{Method}
    & \multirow{2}{*}{\makecell[c]{Sampling \\ steps}}
    & \multirow{2}{*}{$\text{JSD} \downarrow$}
    & \multirow{2}{*}{$\text{MMD} \downarrow$}
    & \multirow{2}{*}{FUD $\downarrow$}
    & \multicolumn{2}{c}{TTCE $\downarrow$}
    & \multicolumn{4}{c}{CTC $\downarrow$}
    & \multirow{2}{*}{ICP Avg. $\downarrow$}
    & \multirow{2}{*}{FPS $\uparrow$} \\
    \cmidrule(lr){6-7}
    \cmidrule(lr){8-11}
    & & & & & 3 & 4 & 1 & 2 & 3 & 4 & & \\
    \midrule
    \multicolumn{4}{l}{\textit{\textbf{HD-Map-conditioned generation}}} \\
    LiDARCrafter~\cite{liu2025lidarcrafter}
    & 256
    & 0.065 & 14.13 & 97.24 & \textbf{1.263} & \textbf{1.989}
    & 0.278 & 0.360 & 0.441 & 0.519
    & 0.120 & 0.127 \\
    LaGen$^*$~\cite{liu2025lagen}
    & 35
    & 0.075 & 19.19 & 50.29 & 1.584 & 2.296
    & \textbf{0.205} & \underline{0.268} & \underline{0.330} & \underline{0.390}
    & \underline{0.071} & 0.402 \\
    Sensor2Sensor$^*$~\cite{wang2026sensor2sensor}
    & 35
    & \underline{0.047} & \underline{5.210} & \underline{37.96} & 1.522 & \underline{2.236}
    & 0.225 & 0.287 & 0.347 & 0.404
    & 0.081 & \underline{0.421} \\
    \textbf{Ours}
    & 4
    & \textbf{0.046} & \textbf{4.390}
    & \textbf{25.01} & \underline{1.491} & 2.255
    & \underline{0.210} & \textbf{0.253} & \textbf{0.293} & \textbf{0.331}
    & \textbf{0.068} & \textbf{11.18} \\
    \midrule
    \multicolumn{4}{l}{\textit{\textbf{Interactive action-conditioned generation}}} \\
    LiDARCrafter~\cite{liu2025lidarcrafter}
    & 256 & 0.137 & 54.383 & 70.38 & \textbf{1.313} & \textbf{2.063} & 0.258 & \underline{0.335} & \underline{0.410} & \underline{0.483} & 0.104 & 0.135\\
    LaGen$^*$~\cite{liu2025lagen}
    & 35 & \underline{0.064} & \underline{12.51} & \underline{42.31} & 1.913 & 2.621 & \underline{0.256} & 0.362 & 0.460 & 0.550 & \underline{0.074} & \underline{0.754} \\
    \textbf{Ours}
    & 4 & \textbf{0.054} & \textbf{7.391} & \textbf{24.17} & \underline{1.517} & \underline{2.280} & \textbf{0.213} & \textbf{0.260} & \textbf{0.305} & \textbf{0.347} & \textbf{0.068} & \textbf{11.30}\\
    \bottomrule
    \end{tabular}    }
\end{table}

\paragraph{HD-Map-Conditioned Generation Results} We report our results in Tab.~\ref{tab:causal-generation}. \methodname{} outperforms all baselines on the statistical and perceptual distribution metrics. It remains competitive on short-term temporal consistency and performs best over longer temporal offsets, indicating that causal distillation preserves the teacher's generative quality without sacrificing long-horizon coherence. The qualitative results in Fig.~\ref{fig:ar_gen_cond} likewise remain structurally stable as the baselines visibly degrade.

\par\medskip\noindent
\begin{minipage}{\linewidth}
\begin{minipage}[t]{0.59\linewidth}
\vspace{0pt}
\paragraph{Action-Conditioned Generation Results} The bottom half of Tab.~\ref{tab:causal-generation} evaluates interactive action-conditioned generation. Given the first ground-truth frame, each method generates 128 additional frames along the ground-truth ego trajectory with no HD-Map conditioning. We omit Sensor2Sensor because it has no native mechanism for injecting ego poses. \methodname{} substantially outperforms the remaining methods across the reported distributional and temporal metrics. The windowed FUD curves in Fig.~\ref{fig:causal-degradation} further show that its quality degrades much more slowly over time, demonstrating that the same causal model can respond to a changing ego trajectory while maintaining a persistent scene.
\end{minipage}\hfill
\begin{minipage}[t]{0.38\linewidth}
\vspace{0pt}\centering
\includegraphics[width=\linewidth]{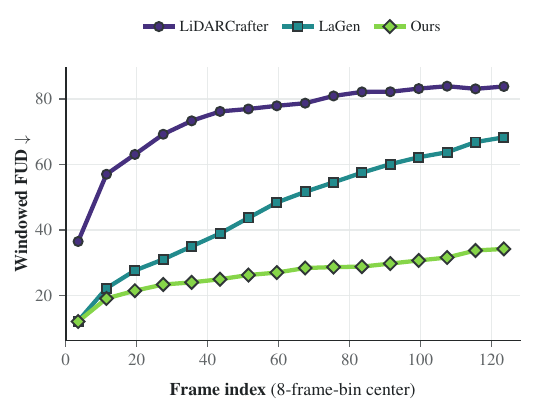}
\captionsetup{hypcap=false}
\captionof{figure}{\textbf{Long-horizon AR rollouts.}}
\label{fig:causal-degradation}
\end{minipage}
\end{minipage}
\par\medskip

\subsection{Point-Cloud Forecasting}
\label{sec:point-cloud-forecasting}
Forecasting introduces more observed evidence than generation: a LiDAR prefix specifies the recent scene state and dynamics, but the future remains unobserved. We evaluate whether the same causal generator can use this prefix to predict future measurements without a task-specific forecasting head. Following \citet{khurana2023point}, we report $L_1$ error, absolute relative error (AbsRel), and inner Chamfer distance (CD) at the one-, two-, and three-second marks.

\begin{figure}[!htbp]
    \centering
    \includegraphics[width=1\linewidth]{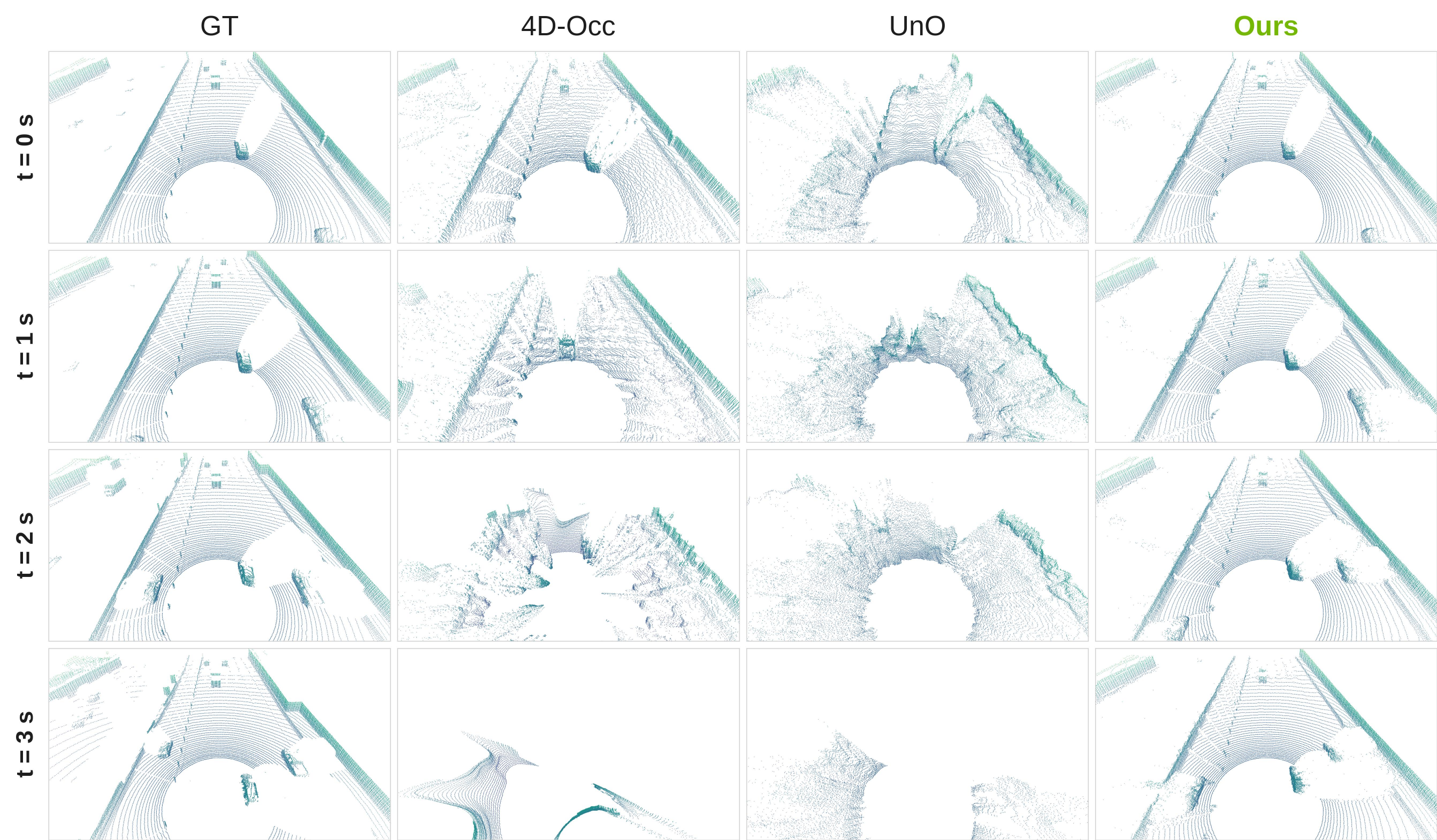}
    \caption{\textbf{Point-cloud forecasting}: \methodname{} preserves dynamic agents and synthesizes plausible geometry as the future viewpoint reveals regions not observed in the input. }
    \label{fig:forecast}
\end{figure}

\paragraph{Settings} Following the point-cloud forecasting protocol of \citet{khurana2023point}, the model predicts a three-second future horizon from three seconds of point-cloud history. We use data sampled at 10\,Hz, resulting in 30 history frames and 30 future frames. We compare against naive raycasting, 4D-Occ~\cite{khurana2023point}, and the state-of-the-art UnO method~\cite{agro2024uno}. Because UnO has not released its code publicly, we reimplement it to the best of our ability based on the details in the paper. For our method, we use a K/V cache size of 21 rather than 9, as we empirically find that it performs better for forecasting. 

\paragraph{Metrics} Following \citet{khurana2023point}, we use $L_1$ error, absolute relative error (AbsRel), and inner Chamfer distance (CD) and evaluate at the one-, two-, and three-second marks. We additionally report aggregate results, computed as the mean of each metric across all 30 predicted frames.

\begin{table}[!htbp]
\centering
\caption{\textbf{Point-cloud forecasting.} $^*$ indicates a reimplementation to the best of our ability.}
\label{tab:forecasting}
\scriptsize
\setlength{\tabcolsep}{3.2pt}
\resizebox{\columnwidth}{!}{%
\begin{tabular}{l *{12}{c}}
\toprule
\multirow{2}{*}{\makecell[c]{Method}} & \multicolumn{3}{c}{1 s} & \multicolumn{3}{c}{2 s}
& \multicolumn{3}{c}{3 s} & \multicolumn{3}{c}{Aggregated} \\
\cmidrule(lr){2-4}\cmidrule(lr){5-7}\cmidrule(lr){8-10}\cmidrule(lr){11-13}
& $L_1$~$\downarrow$ & AbsRel~$\downarrow$ & CD~$\downarrow$
& $L_1$~$\downarrow$ & AbsRel~$\downarrow$ & CD~$\downarrow$
& $L_1$~$\downarrow$ & AbsRel~$\downarrow$ & CD~$\downarrow$
& $L_1$~$\downarrow$ & AbsRel~$\downarrow$ & CD~$\downarrow$\\
\midrule
Raycasting~\cite{khurana2023point}
& 2.461 & 0.171 & 1.584
& 3.161 & 0.217 & 3.343
& 3.768 & 0.251 & 6.049
& 2.807 & 0.193 & 2.739\\
4D-Occ~\cite{khurana2023point}
& 2.146 & 0.113 & 2.173
& 2.602 & 0.139 & 3.230
& 3.052 & 0.170 & 4.575
& 2.369 & \underline{0.126} & 2.924\\
UnO$^*$~\cite{agro2024uno}
& \underline{2.040} & \underline{0.112} & \underline{1.000}
& \underline{2.300} & \underline{0.130} & \underline{1.117}
& \underline{2.654} & \underline{0.165} & \underline{1.440}
& \underline{2.208} & \underline{0.126} & \underline{1.070} \\
\midrule
\textbf{Ours}
& \textbf{1.482} & \textbf{0.090} & \textbf{0.485}
& \textbf{2.104} & \textbf{0.126} & \textbf{0.805}
& \textbf{2.554} & \textbf{0.156} & \textbf{1.328}
& \textbf{1.768} & \textbf{0.108} & \textbf{0.676}\\
\bottomrule
\end{tabular}%
}
\end{table}

\paragraph{Results} Tab.~\ref{tab:forecasting} shows that \methodname{} outperforms the task-specific baselines on all reported metrics across all forecast horizons. The visualizations in Fig.~\ref{fig:forecast} show plausible future motion while preserving the shapes of traffic participants. Naive raycasting struggles when the future viewpoint reveals regions absent from the input history, as in the freeway example; the shared generative prior can instead complete plausible scene structure. These results position forecasting as an intermediate inference regime of the same world model rather than a separate prediction architecture.

\subsection{Generalizable LiDAR Resimulation}
\label{sec:lidar-resimulation}

Resimulation is the most observation-constrained regime. Sparse sweeps identify a particular captured world, and the PSC projects their accumulated geometry into each requested sensor viewpoint. This setting tests whether a cross-scene generative model can recover the fidelity and viewpoint control associated with reconstruction-based simulation without fitting a representation to each test scene.

\paragraph{Settings} Given sparse observed sweeps of a captured scene, each method synthesizes LiDAR observations at requested sensor poses. Quantitative evaluation uses held-out frames along the recorded sensor trajectory, where ground truth is available. We report results on the 202-sequence Waymo validation split and its top-20 travel-distance subset. Following \citet{storm} and \citet{chen2026dggt}, we use non-overlapping 20-frame chunks with frames 0, 4, 9, and 14 as input. The 16 remaining frames comprise 11 interpolation frames (frames 0--14 excluding the four inputs) and five extrapolation frames (frames 15--19). Aggregate metrics cover all 16 evaluated frames.

\paragraph{Travel-Distance Subset}
We rank all 202 Waymo validation sequences by total ego path length, computed as the sum of 3D translation distances between consecutive sweep-start LiDAR poses over the full sequence. We retain the 20 longest paths and divide each into nine non-overlapping 20-frame windows, yielding 180 windows. We report the sequence IDs in Tab.~\ref{tab:dense-top20-sequences}.
\begin{figure}[H]
    \centering
    \includegraphics[width=1\linewidth]{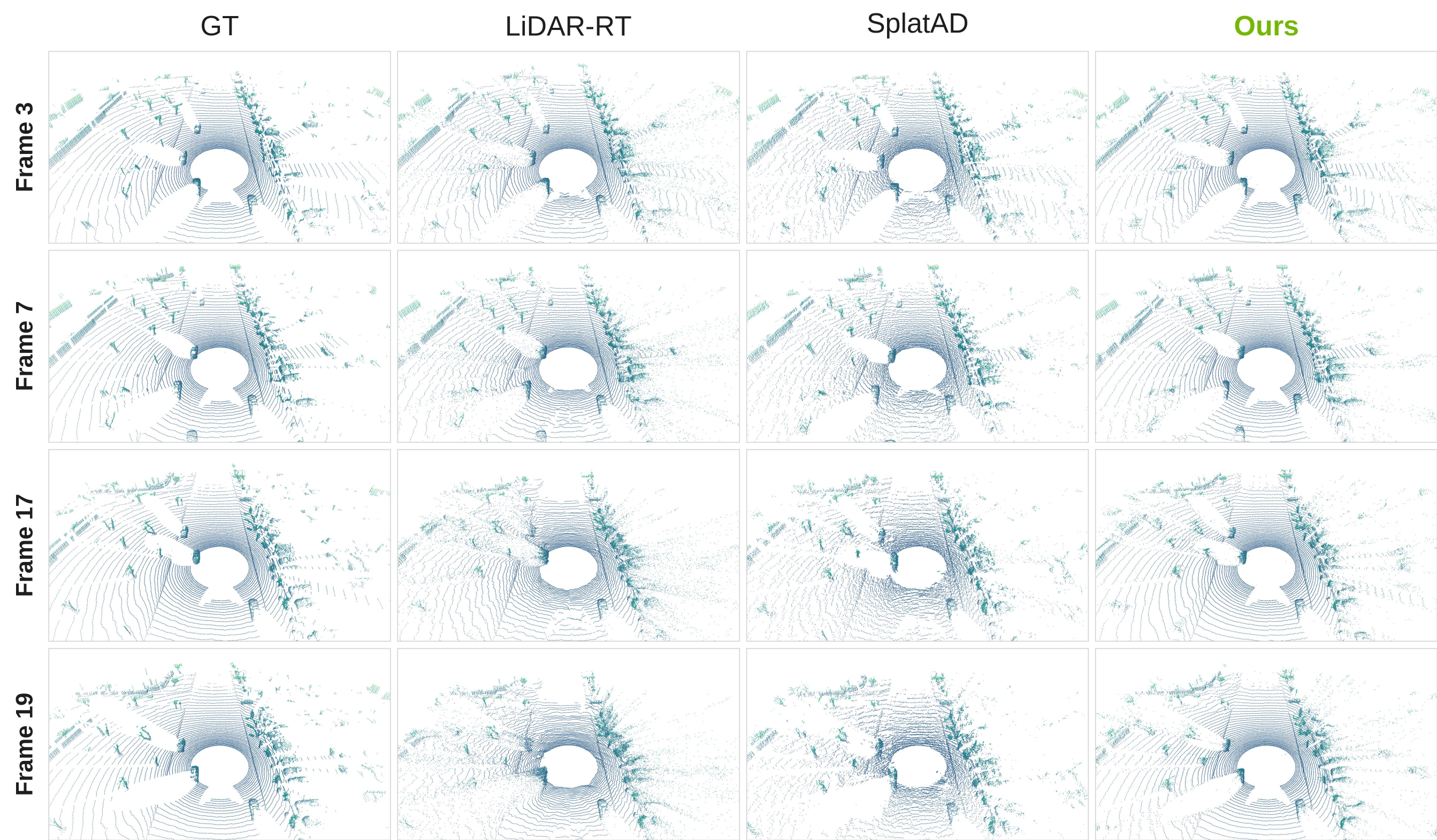}
    \caption{\textbf{Generalizable LiDAR resimulation}: the causal \methodname{} checkpoint achieves strong resimulation fidelity in both the interpolation setting (frames 3 and 7) and the extrapolation setting (frames 17 and 19). It requires no per-scene optimization and renders in real time.}
    \label{fig:resim}
\end{figure}

\begin{table}[!htbp]
\centering
\caption{\textbf{Top-20 travel-distance validation sequences.} Sequences are ranked by total ego path length over the full log.}
\label{tab:dense-top20-sequences}
\scriptsize
\setlength{\tabcolsep}{5pt}
\begin{tabular}{rllrll}
\toprule
Rank & Sequence ID && Rank & Sequence ID \\
\midrule
1  & \texttt{\detokenize{2506799708748258165_6455_000_6475_000}} &&
11 & \texttt{\detokenize{16767575238225610271_5185_000_5205_000}} \\
2  & \texttt{\detokenize{5372281728627437618_2005_000_2025_000}} &&
12 & \texttt{\detokenize{13573359675885893802_1985_970_2005_970}} \\
3  & \texttt{\detokenize{6491418762940479413_6520_000_6540_000}} &&
13 & \texttt{\detokenize{271338158136329280_2541_070_2561_070}} \\
4  & \texttt{\detokenize{15396462829361334065_4265_000_4285_000}} &&
14 & \texttt{\detokenize{10203656353524179475_7625_000_7645_000}} \\
5  & \texttt{\detokenize{18305329035161925340_4466_730_4486_730}} &&
15 & \texttt{\detokenize{10335539493577748957_1372_870_1392_870}} \\
6  & \texttt{\detokenize{13415985003725220451_6163_000_6183_000}} &&
16 & \texttt{\detokenize{1457696187335927618_595_027_615_027}} \\
7  & \texttt{\detokenize{9164052963393400298_4692_970_4712_970}} &&
17 & \texttt{\detokenize{2105808889850693535_2295_720_2315_720}} \\
8  & \texttt{\detokenize{10837554759555844344_6525_000_6545_000}} &&
18 & \texttt{\detokenize{11048712972908676520_545_000_565_000}} \\
9  & \texttt{\detokenize{4490196167747784364_616_569_636_569}} &&
19 & \texttt{\detokenize{14165166478774180053_1786_000_1806_000}} \\
10 & \texttt{\detokenize{15096340672898807711_3765_000_3785_000}} &&
20 & \texttt{\detokenize{7163140554846378423_2717_820_2737_820}} \\
\bottomrule
\end{tabular}

\end{table}

\paragraph{Baselines} We compare with SplatAD~\cite{hess2025splatad}, LiDAR4D~\cite{zheng2024lidar4d}, and LiDAR-RT~\cite{zhou2025lidarrt}, using each method's released codebase. These baselines optimize a separate scene representation for every test sequence. In contrast, \methodname{} uses one checkpoint across all scenes and requires no test-time optimization, making it a generalizable LiDAR resimulation model.

\paragraph{Metrics} We report range $L_1$ error and Chamfer distance (CD) for interpolation, extrapolation, and their aggregate in Tab.~\ref{tab:resimulation}. We additionally report per-scene optimization time and whether rendering operates in real time.

\begin{table}[!htbp]
\centering
\caption{\textbf{Generalizable LiDAR resimulation.} The causal student is our default real-time model; the bidirectional teacher is included as a quality reference. Per-scene optimization time is measured in minutes.}
\label{tab:resimulation}
\scriptsize
\setlength{\tabcolsep}{3.2pt}
\resizebox{\columnwidth}{!}{%
\begin{tabular}{l *{14}{c}}
\toprule
\multirow{3}{*}{\makecell[c]{Method}}
& \multicolumn{6}{c}{All-202}
& \multicolumn{6}{c}{Top-20 Travel Distance}
& \multirow{3}{*}{\makecell[c]{Optim.\\time}$\downarrow$}
& \multirow{3}{*}{\makecell[c]{Real-time\\render}} \\
\cmidrule(lr){2-7}\cmidrule(lr){8-13}
& \multicolumn{2}{c}{Interpolation}
& \multicolumn{2}{c}{Extrapolation}
& \multicolumn{2}{c}{Aggregate}
& \multicolumn{2}{c}{Interpolation}
& \multicolumn{2}{c}{Extrapolation}
& \multicolumn{2}{c}{Aggregate} \\
\cmidrule(lr){2-3}\cmidrule(lr){4-5}\cmidrule(lr){6-7}
\cmidrule(lr){8-9}\cmidrule(lr){10-11}\cmidrule(lr){12-13}
& $L_1\downarrow$ & CD$\downarrow$
& $L_1\downarrow$ & CD$\downarrow$
& $L_1\downarrow$ & CD$\downarrow$
& $L_1\downarrow$ & CD$\downarrow$
& $L_1\downarrow$ & CD$\downarrow$
& $L_1\downarrow$ & CD$\downarrow$ \\
\midrule
LiDAR4D~\cite{zheng2024lidar4d}
& 1.129 & 0.188
& 1.881 & 0.460
& 1.364 & 0.273
& 1.999 & 0.998
& 3.284 & 1.769
& 2.401 & 1.239
& 89.8 & \textcolor{red!75!black}{\(\times\)} \\
LiDAR-RT~\cite{zhou2025lidarrt}
& \textbf{1.089} & \underline{0.182}
& \underline{1.593} & 0.418
& \underline{1.247} & \underline{0.256}
& 1.905 & 0.444
& 3.094 & 1.292
& 2.277 & 0.709
& 76.3 & \textcolor{green!60!black}{\checkmark} \\
SplatAD~\cite{hess2025splatad}
& 1.799 & 0.265
& 2.325 & 0.408
& 1.964 & 0.309
& 3.125 & 0.357
& 4.654 & 0.947
& 3.603 & 0.542
& \underline{15.0} & \textcolor{green!60!black}{\checkmark} \\
\midrule
\textbf{Ours (bidir)}
& \underline{1.124} & \textbf{0.169}
& \textbf{1.376} & \textbf{0.270}
& \textbf{1.202} & \textbf{0.200}
& \textbf{1.399} & \textbf{0.184}
& \textbf{1.838} & \textbf{0.354}
& \textbf{1.536} & \textbf{0.237}
& \textbf{0.00} & \textcolor{red!75!black}{\(\times\)} \\
\textbf{Ours}
& 1.279 & 0.235
& 1.682 & \underline{0.397}
& 1.405 & 0.285
& \underline{1.593} & \underline{0.260}
& \underline{2.215} & \underline{0.370}
& \underline{1.787} & \underline{0.294}
& \textbf{0.00} & \textcolor{green!60!black}{\checkmark} \\
\bottomrule
\end{tabular}%
}
\vspace{-2em}
\end{table}

\paragraph{Results} Tab.~\ref{tab:resimulation} shows that despite requiring no test-time optimization, our causal student achieves competitive performance with the per-scene reconstruction baselines on the full validation cohort and outperforms them on every reported quality metric in the top-20 travel-distance subset. The bidirectional teacher provides a quality reference and achieves the best fidelity across nearly all reported metrics, with LiDAR-RT attaining the lowest interpolation $L_1$ on All-202. The qualitative results in Fig.~\ref{fig:resim} further show that the causal checkpoint preserves captured scene structure in both interpolation and extrapolation views. To our knowledge, \methodname{} is the first real-time generalizable LiDAR resimulation model of this capability. Together with generation and forecasting, these results show that one conditional generator spans generative and reconstruction-grounded sensor simulation.

\subsection{Intensity Modeling}
\label{sec:intensity-modeling}
\begin{wraptable}{r}{0.42\textwidth}
  \centering
  \small
  \vspace{-2em}
  \caption{\textbf{Generalizable LiDAR resimulation intensity RMSE on the top-20 travel-distance subset.} Ours denotes the intensity-fine-tuned bidirectional teacher. Lower is better.}
  \label{tab:resim-intensity}
  \vspace{-0.5em}
  \begin{tabular}{lccc}
    \toprule
    Method & Interp. & Extrap. & Agg. \\
    \midrule
    LiDAR4D & \textbf{0.077} & \underline{0.094} & \textbf{0.082} \\
    LiDAR-RT & 0.083 & 0.100 & 0.089 \\
    SplatAD & 0.115 & 0.134 & 0.121 \\
    \midrule
    \textbf{Ours}     & \underline{0.078} & \textbf{0.091} & \textbf{0.082} \\
    \bottomrule
  \end{tabular}
\end{wraptable}
Sensor simulation must capture not only range geometry but also auxiliary measurements used by downstream perception. We therefore extend the model to LiDAR intensity, which is particularly important for tasks such as semantic segmentation. We fine-tune the tokenizer to reconstruct intensity and subsequently fine-tune the bidirectional teacher on the combined range-intensity latent representation. Using the resimulation protocol for the top-20 travel-distance subset from Sec.~\ref{sec:lidar-resimulation}, our intensity predictions remain competitive with those of the state-of-the-art reconstruction baselines (Tab.~\ref{tab:resim-intensity}).

\subsection{Generalization Across Sensors and Datasets}
\label{sec:cross-dataset}

A general LiDAR representation should not be tied to the scanning pattern of one sensor or the distribution of one dataset. We therefore evaluate whether the attention-based tokenizer and shared generative interface can be adapted jointly across heterogeneous LiDAR configurations.

\paragraph{Settings} Most existing LiDAR tokenizers target a single sensor configuration. Benefiting from its purely Transformer-based architecture and lack of temporal downsampling, our tokenizer can instead adapt to different scanning patterns, resolutions, and operating frequencies within one checkpoint. We jointly fine-tune the Waymo-trained tokenizer on 32-beam nuScenes~\cite{nuscenes}, 64-beam Waymo, and 128-beam NVIDIA PAI~\cite{nvidiapai}. We then fine-tune the bidirectional teacher on the same mixture using an additional sensor embedding that identifies each dataset's LiDAR configuration, while retaining the model architecture and training recipe. For cross-dataset intensity generation, we apply the same procedure to a tokenizer and DiT trained on both range and intensity.

\begin{figure}[!htbp]
    \centering
    \includegraphics[width=1\linewidth]{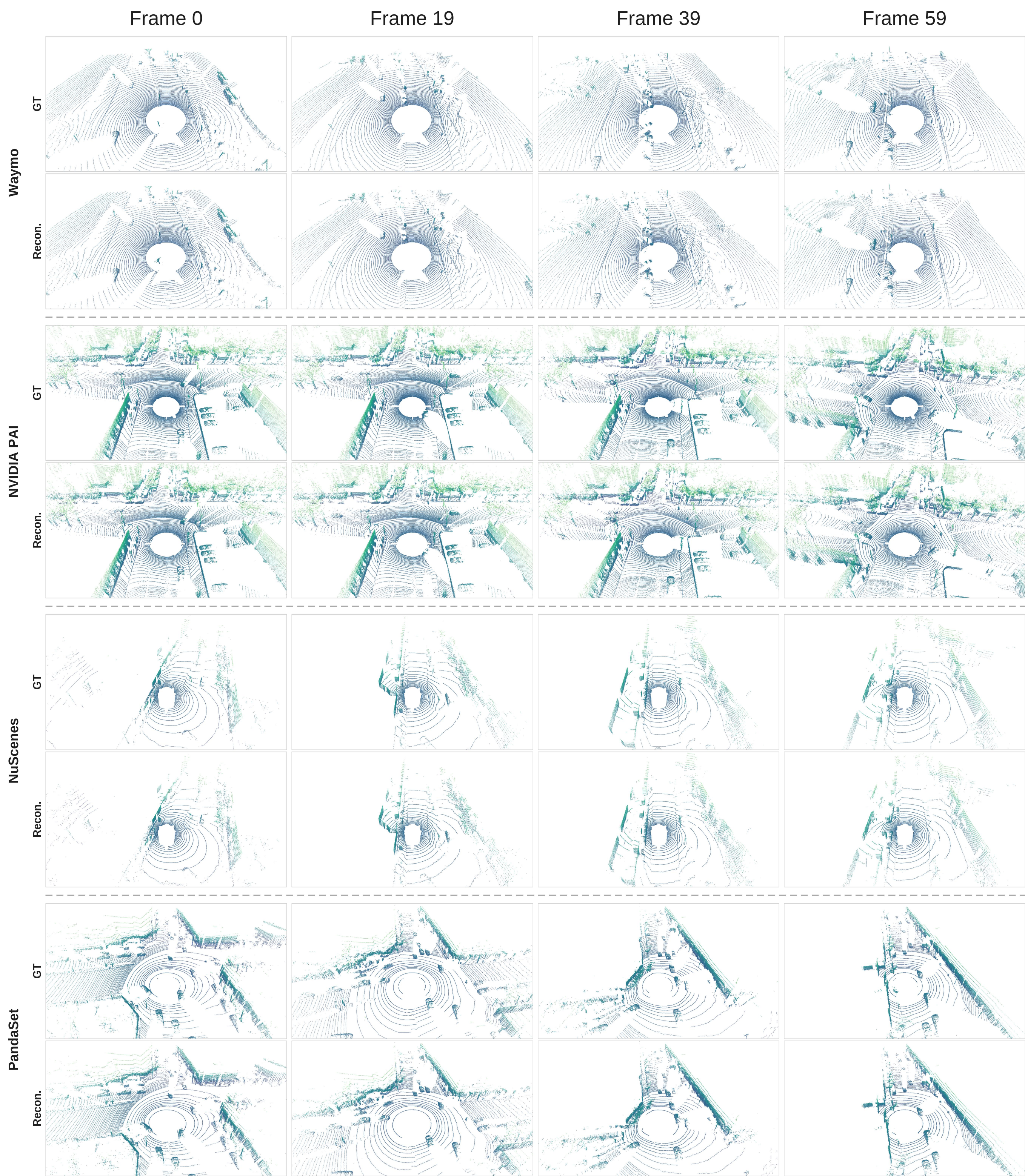}
    \caption{\textbf{Cross-dataset reconstruction.} One tokenizer handles 32-, 64-, and 128-beam LiDAR configurations and generalizes to PandaSet, which is not included in training.}
    \label{fig:cross-dataset-recon}
\end{figure}

\begin{wraptable}{r}{0.42\textwidth}
  \centering
  \small
  \setlength{\tabcolsep}{2.5pt}
  \caption{\textbf{Universal vs.\ specialist tokenizer reconstruction.}
  Reconstruction is evaluated on matched 21-frame sequences. Metrics are measured in meters; lower is better.}
  \label{tab:universal-vs-specialist}
  \vspace{-0.5em}
  \begin{tabular}{@{}llccc@{}}
    \toprule
    Metric & Model & nuScenes & Waymo & NVIDIA PAI \\
    \midrule
    \multirow{2}{*}{$L_1 \downarrow$}
      & Specialist & 0.087 & 0.056 & 0.286 \\
      & \textbf{Universal}
      & \textbf{0.065} & \textbf{0.037} & \textbf{0.087} \\
    \midrule
    \multirow{2}{*}{CD $\downarrow$}
      & Specialist & \textbf{0.557} & 0.718 & 1.240 \\
      & \textbf{Universal}
      & 0.568 & \textbf{0.710} & \textbf{1.135} \\
    \bottomrule
  \end{tabular}
\end{wraptable}
\paragraph{Reconstruction Results} Figure~\ref{fig:cross-dataset-recon} shows reconstructions from the universal tokenizer for all three training sensor configurations, as well as its generalization to the unseen PandaSet~\cite{pandaset} scanning pattern. To isolate the benefit of cross-dataset training, we compare its reconstruction quality on each dataset with that of a specialist tokenizer trained only on that dataset. For a fair comparison, all models use the same architecture and number of training iterations and are evaluated on matched 21-frame sequences; the tokenizer comparison in Tab.~\ref{tab:vae-comparison} instead uses nine-frame sequences. Table~\ref{tab:universal-vs-specialist} indicates that joint training across datasets improves reconstruction quality over each specialist.

\begin{figure}[!htbp]
    \centering
    \includegraphics[width=1\linewidth]{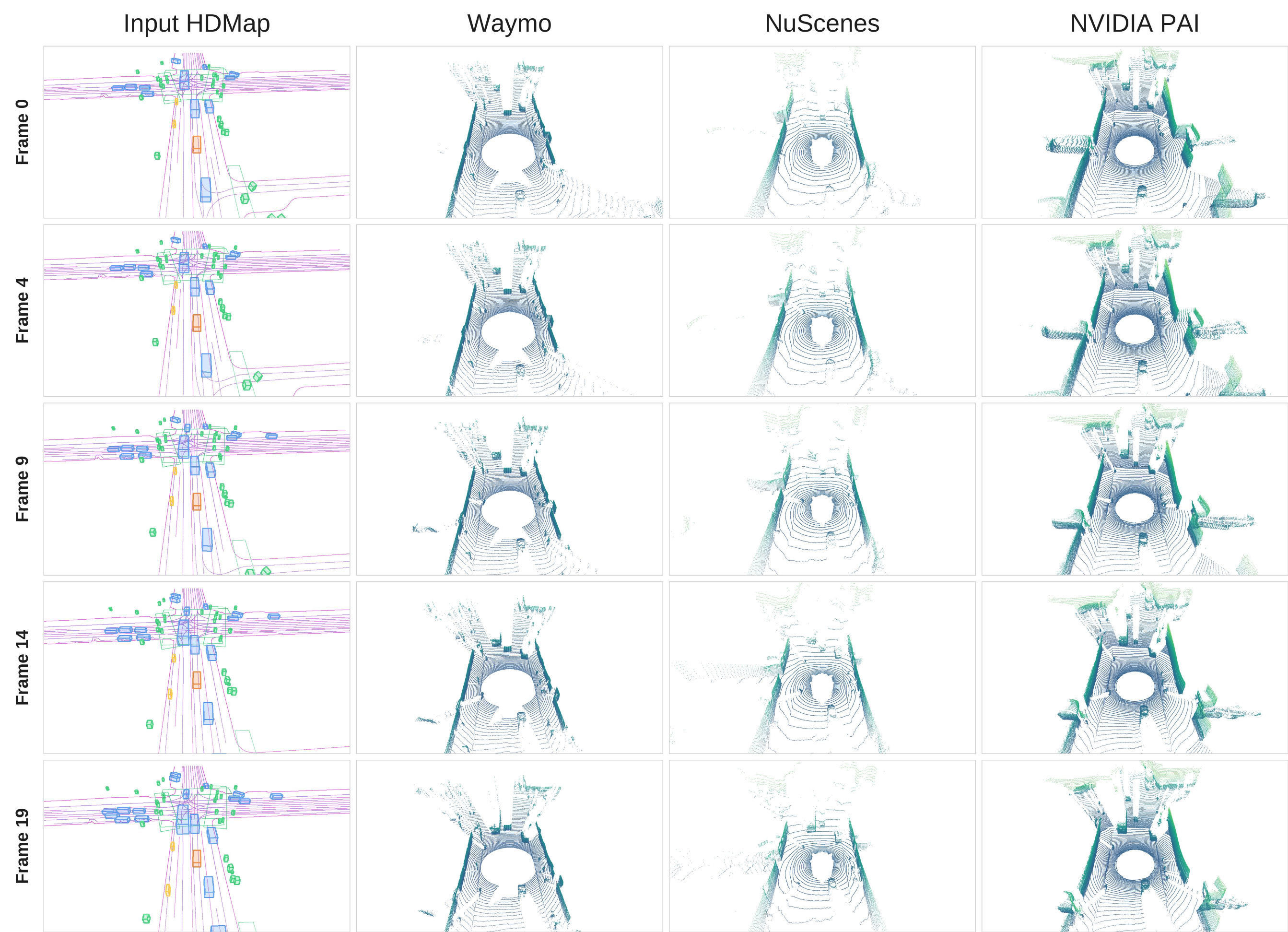}
    \caption{\textbf{Cross-dataset generation.} By changing the sensor embedding, our fine-tuned bidirectional teacher generates distinct LiDAR scanning patterns from the same Waymo HD-Map input.}
    \label{fig:cross-dataset-gen}
\end{figure}

\begin{figure}[!htbp]
    \centering
    \includegraphics[width=1\linewidth]{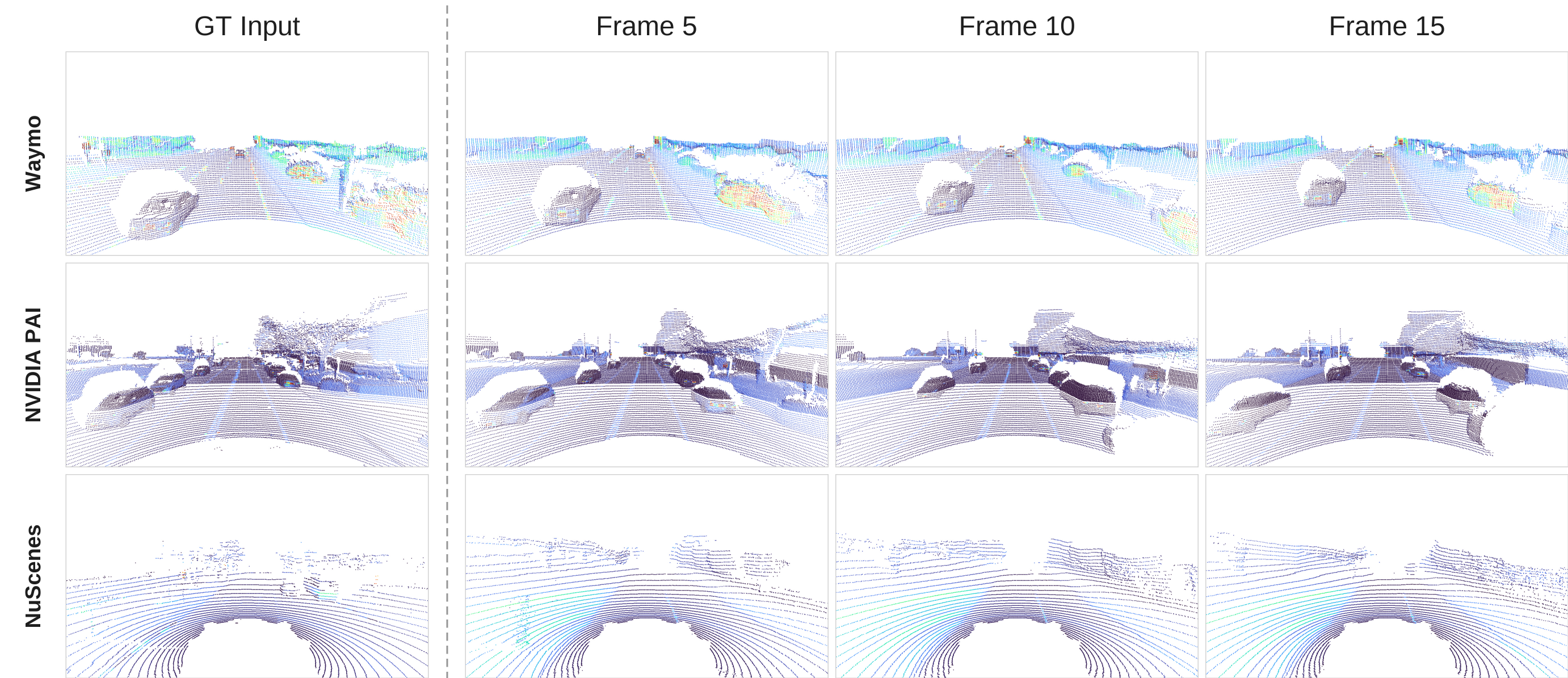}
    \caption{\textbf{Cross-dataset first-frame-conditioned generation with intensity.} The fine-tuned bidirectional teacher produces high-quality cross-dataset generation with intensity in a single model, enabled by the universal tokenizer. Each sequence is conditioned on its first ground-truth LiDAR frame.}
    \label{fig:cross-dataset-intensity}
\end{figure}

\paragraph{Generation Results} The jointly trained bidirectional model also generates the distinct nuScenes, NVIDIA PAI, and Waymo LiDAR patterns from the same Waymo HD-Map input (Fig.~\ref{fig:cross-dataset-gen}). Moreover, the cross-dataset intensity model produces future frames with plausible intensity profiles, including high-intensity returns from retroreflective surfaces (Fig.~\ref{fig:cross-dataset-intensity}).

\subsection{Ablation}
\label{sec:ablation}

We finally isolate the design choices behind the shared representation and persistent rollout. The tokenizer ablation asks whether attention is necessary for LiDAR geometry, while the system ablations separate short-term temporal context from long-term spatial memory and examine how conservatively generated geometry should enter the PSC. We also compare the unified teacher with budget-matched task-specific specialists.

\begin{figure}[!htbp]
    \centering
    \begin{minipage}[t]{0.67\textwidth}
        \vspace{0pt}
        \centering
        \captionof{table}{\textbf{Tokenizer architecture ablation on the Waymo dataset.} We report representative reconstruction, generation, temporal-consistency, and efficiency metrics, demonstrating the superior performance of our attention backbone compared with the conventional convolutional backbone.}
        \label{tab:vae-ablation}
        \scriptsize
        \setlength{\tabcolsep}{2.2pt}
        \resizebox{\linewidth}{!}{%
        \begin{tabular}{l ccc c cc c}
        \toprule
        \multirow{2}{*}{Backbone}
        & \multicolumn{3}{c}{Recon.}
        & \multicolumn{3}{c}{Generation}
        & \multirow{2}{*}{\makecell[c]{TFLOPs\\/frame$\downarrow$}} \\
        \cmidrule(lr){2-4}\cmidrule(lr){5-7}
        & $L_1$ $\downarrow$
        & CD $\downarrow$
        & AbsRel~$\downarrow$
        & FUD $\downarrow$
        & CTC@4 $\downarrow$
        & TTCE@4 $\downarrow$
        & \\
        \midrule
        Conv.
        & 0.297 & 0.278 & 0.015
        & 26.54 & 0.341 & 2.496 & 0.575 \\
        Attn.
        & \textbf{0.072} & \textbf{0.115} & \textbf{0.005}
        & \textbf{18.17} & \textbf{0.339} & \textbf{2.463}
        & \textbf{0.250} \\
        \bottomrule
        \end{tabular}%
        }
    \end{minipage}%
    \hfill
    \begin{minipage}[t]{0.3\textwidth}
        \vspace{0pt}
        \centering
        \includegraphics[width=\linewidth]{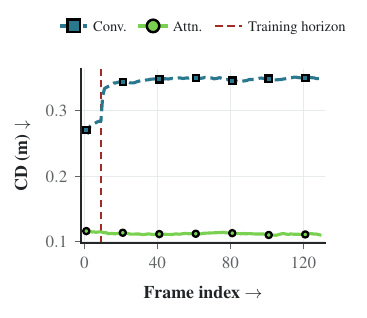}
        \vspace{-2.5em}
        \captionof{figure}{\textbf{Long-sequence recon.
        }
        }
        \label{fig:sup-vae-sequence-generalization}
    \end{minipage}
\end{figure}

\paragraph{LiDAR-Native Tokenizer Architecture} We isolate the effect of spatial neighborhood attention by comparing our tokenizer against a convolutional tokenizer at the same $8{\times}8{\times}1$ compression rate and latent dimensionality. Each tokenizer is paired with the same DiT architecture and training protocol. As shown in Tab.~\ref{tab:vae-ablation}, neighborhood attention improves reconstruction across all metrics and yields better long-horizon temporal consistency while reducing tokenizer compute by $57\%$ relative to the convolutional baseline. Figure~\ref{fig:sup-vae-sequence-generalization} further shows that the attention-based tokenizer reconstructs sequences more stably beyond its training horizon. Both backbones are trained on nine-frame sequences and evaluated on 129-frame sequences from 202 Waymo validation clips using EMA weights.

\begin{figure}[!htbp]
    \centering
    \begin{minipage}[t]{0.48\textwidth}
        \vspace{0pt}
        \centering
        \includegraphics[width=\linewidth]{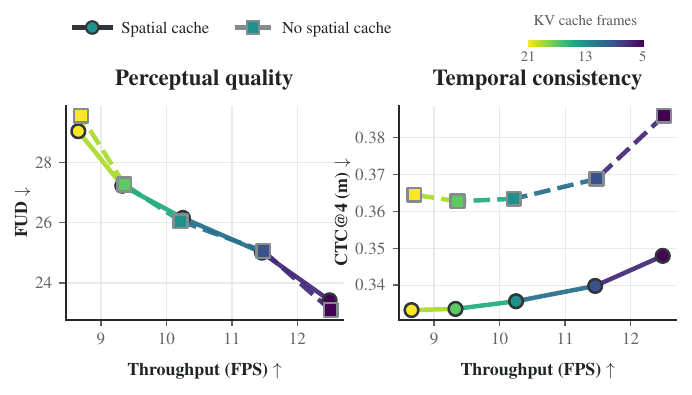}
        \vspace{-2em}
        \captionof{figure}{\textbf{Spatial and K/V cache size ablation.} }
        \label{fig:spatial-cache-kv}
    \end{minipage}%
    \hfill
    \begin{minipage}[t]{0.48\textwidth}
        \vspace{0pt}
        \centering
        \includegraphics[width=\linewidth]{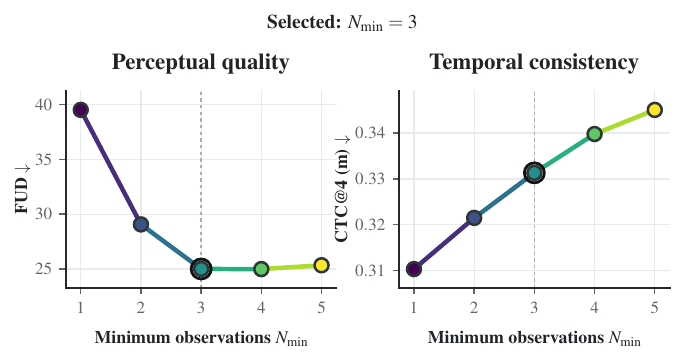}
        \captionof{figure}{\textbf{Spatial-cache confirmation ablation.} }
        \label{fig:spatial-cache-minobs}
    \end{minipage}
\end{figure}

\paragraph{Spatial Cache and Temporal K/V Cache}
We jointly vary the temporal K/V cache length over $\{5,9,13,17,21\}$ frames and either enable the PSC condition or replace it with the learned null condition. We fix $N_{\min}=3$ for this sweep. Perceptual and temporal metrics use all 202 Waymo validation logs under the 129-frame generation protocol, while steady-state throughput is measured on ten matched clips on an H100 GPU. Figure~\ref{fig:spatial-cache-kv} shows the resulting quality--throughput trade-off. Removing the spatial cache substantially worsens temporal consistency while having little effect on FUD. Increasing the K/V cache size improves temporal consistency but reduces throughput and fidelity. We therefore use a nine-frame K/V cache together with the spatial cache, which provides a favorable balance among fidelity, temporal consistency, and real-time throughput.

\paragraph{Spatial-Cache Confirmation} Using the same 202 validation logs and a fixed nine-frame K/V cache, we vary the observation threshold $N_{\min}\in\{1,2,3,4,5\}$ required to promote a candidate voxel into the PSC. As shown in Fig.~\ref{fig:spatial-cache-minobs}, directly registering the latest generated sweep (i.e., $N_{\min}=1$) produces worse fidelity despite yielding the best temporal-consistency metrics. We choose $N_{\min}=3$ to balance fidelity and temporal consistency, and use this setting for the main causal-generation results in Tab.~\ref{tab:causal-generation}.

\Needspace{14\baselineskip}
\begin{wraptable}{r}{0.48\textwidth}
    \centering
    \vspace{-1em}
    \caption{\textbf{Unified versus specialists.}  MMD values are multiplied by 10$^5$.}
    \label{tab:unified-specialized}
    \footnotesize
    \setlength{\tabcolsep}{2pt}
    \renewcommand{\arraystretch}{1.1}
    \begin{tabular*}{\linewidth}{@{\extracolsep{\fill}}lccccc@{}}
        \toprule
        \multirow{2}{*}{Method}
        & \multicolumn{3}{c}{Generation}
        & \multicolumn{2}{c}{Resimulation} \\
        \cmidrule(lr){2-4}\cmidrule(lr){5-6}
        & FUD~$\downarrow$ & \makecell{MMD~$\downarrow$} & ICP Avg.~$\downarrow$
        & $L_1$~$\downarrow$ & CD~$\downarrow$ \\
        \midrule
        Specialist & \textbf{14.67} & 6.950 & 0.047 & \textbf{1.380} & \textbf{0.235} \\
        Unified & 16.05 & \textbf{6.843} & \textbf{0.044} & 1.536 & 0.237 \\
        \bottomrule
    \end{tabular*}
\end{wraptable}
\paragraph{Unified versus Task-Specific Specialists}
We compare our unified bidirectional model with two task-specific specialists for generation and resimulation. Each specialist is trained with the same budget as the unified model. For HD-Map-conditioned generation, both models use 35 sampling steps and generate 21-frame clips over all 202 Waymo validation sequences. Resimulation follows Sec.~\ref{sec:lidar-resimulation}, with metrics aggregated over the top-20 travel-distance subset. The specialist row in Tab.~\ref{tab:unified-specialized} combines the results of two separately trained models; the unified row uses one checkpoint for both tasks. Unified training retains competitive performance across both paradigms: it improves generation MMD and ICP Avg., while the specialists achieve lower FUD and resimulation $L_1$ and CD. This enables a single checkpoint to cover both generation and resimulation with modest trade-offs on the reported metrics.

\FloatBarrier

\section{Limitations and Future Work}
\label{sec:limitations}
Currently, \methodname{} models a single spinning LiDAR. Future work
could extend the framework to solid-state LiDAR and joint generation
for multi-LiDAR rigs, leveraging datasets such as the Waymo Open
Dataset~\citep{sun2020waymo} and Argoverse~2~\citep{Argoverse2}.
Second, \methodname{} still relies on annotated object poses for dynamic
resimulation during both training and inference, limiting its scalability
to unlabeled scenes. While recent RGB-based methods have explored
generalizable reconstruction without such object-level
supervision~\citep{storm, wang2025flux4d, lu2025drivingrecon},
annotation-free generalizable resimulation remains an important
open challenge for LiDAR.
Finally, \methodname{} focuses exclusively on LiDAR. Extending the
framework to jointly generate synchronized observations across sensor
modalities, such as cameras and LiDAR, is a promising direction toward
more comprehensive autonomous-driving simulation, following recent
advances~\citep{cosmosdrivedreams, wang2026sensor2sensor}.

\section{Conclusion}
\label{sec:conclusion}

We presented \methodname{}, a real-time LiDAR world model that unifies generative and reconstruction-grounded sensor simulation. Rather than choosing between a distributional generator and a scene-specific reconstruction, \methodname{} expresses both within one conditional generative framework: scene-level conditions guide new-world generation, an observed LiDAR prefix enables point-cloud forecasting, and a persistent spatial cache anchors generalizable resimulation to captured geometry without per-scene optimization. A universal LiDAR tokenizer provides a shared latent space across sensor configurations, while a bidirectional diffusion teacher learns the common conditioning interface.

We distill this teacher into a four-step causal student that combines a rolling temporal K/V cache for recent dynamics with a world-aligned spatial cache for persistent geometry. The resulting causal checkpoint operates at or above the native 10-Hz LiDAR rate and achieves state-of-the-art performance on most reported metrics across generation, forecasting, and resimulation. The tokenizer additionally supports 32-, 64-, and 128-beam LiDAR and generalizes to an unseen scanning pattern. Together, these results show that changing the available scene evidence, rather than the model architecture, can move one LiDAR world model between generative simulation and reconstruction-grounded resimulation. Although we do not yet evaluate a complete closed-loop system, the model's interactive action conditioning and requested-pose resimulation provide key capabilities toward that goal.

\clearpage
{
\small

\bibliographystyle{unsrtnat}
\bibliography{references}

\begin{thebibliography}{66}
\providecommand{\natexlab}[1]{#1}
\providecommand{\url}[1]{\texttt{#1}}
\expandafter\ifx\csname urlstyle\endcsname\relax
  \providecommand{\doi}[1]{doi: #1}\else
  \providecommand{\doi}{doi: \begingroup \urlstyle{rm}\Url}\fi

\bibitem[{NVIDIA}(2024)]{nvidia2024nurec}
{NVIDIA}.
\newblock {NVIDIA Omniverse NuRec}: Neural reconstruction for autonomous
  vehicle simulation.
\newblock \url{https://developer.nvidia.com/omniverse/nurec}, 2024.

\bibitem[Basant et~al.(2026)Basant, Kar, Paschalidou, Garcia~Cobo, Turki, Ling,
  Seo, Wang, Lucas, Wu, Lorraine, Gao, He, Tothova, Xie, Tyszkiewicz, Wu,
  de~Lutio, Li, Fidler, Kim, Shen, Cao, Pfaff, Lew, Ren, Lu, Gojcic, and
  Wang]{nvidia2026omnidreams}
Aarti Basant, Amlan Kar, Despoina Paschalidou, Guillermo Garcia~Cobo, Haithem
  Turki, Huan Ling, Jaewoo Seo, Jialiang Wang, James Lucas, Jay Wu, Jonathan
  Lorraine, Jun Gao, Kai He, Katarina Tothova, Kevin Xie, Michal Tyszkiewicz,
  Qi~Wu, Riccardo de~Lutio, Ruilong Li, Sanja Fidler, Seung~Wook Kim, Tianchang
  Shen, Tianshi Cao, Tobias Pfaff, William Lew, Xuanchi Ren, Yifan Lu, Zan
  Gojcic, and Zian Wang.
\newblock {OmniDreams}: Real-time generative closed-loop autonomous vehicle
  simulation built on {NVIDIA Cosmos}.
\newblock Technical report, 2026.

\bibitem[Wu et~al.(2024)Wu, Zuo, Leutenegger, Litany, Schindler, and
  Huang]{dynfl}
Hanfeng Wu, Xingxing Zuo, Stefan Leutenegger, Or~Litany, Konrad Schindler, and
  Shengyu Huang.
\newblock Dynamic {LiDAR} re-simulation using compositional neural fields.
\newblock In \emph{Proceedings of the IEEE/CVF Conference on Computer Vision
  and Pattern Recognition (CVPR)}, 2024.

\bibitem[Huang et~al.(2023)Huang, Gojcic, Wang, Williams, Kasten, Fidler,
  Schindler, and Litany]{huang2023nfl}
Shengyu Huang, Zan Gojcic, Zian Wang, Francis Williams, Yoni Kasten, Sanja
  Fidler, Konrad Schindler, and Or~Litany.
\newblock Neural {LiDAR} fields for novel view synthesis.
\newblock In \emph{Proceedings of the IEEE/CVF International Conference on
  Computer Vision (ICCV)}, 2023.

\bibitem[Zheng et~al.(2024)Zheng, Lu, Xue, Chen, and Jiang]{zheng2024lidar4d}
Zehan Zheng, Fan Lu, Weiyi Xue, Guang Chen, and Changjun Jiang.
\newblock {LiDAR4D}: Dynamic neural fields for novel space-time view {LiDAR}
  synthesis.
\newblock In \emph{Proceedings of the IEEE/CVF Conference on Computer Vision
  and Pattern Recognition (CVPR)}, 2024.

\bibitem[Zhou et~al.(2025{\natexlab{a}})Zhou, Fu, Peng, Yan, Zhang, Chen, Xia,
  and Zhou]{zhou2025lidarrt}
Chenxu Zhou, Lvchang Fu, Sida Peng, Yunzhi Yan, Zhanhua Zhang, Yong Chen,
  Jiazhi Xia, and Xiaowei Zhou.
\newblock {LiDAR-RT}: Gaussian-based ray tracing for dynamic {LiDAR}
  re-simulation.
\newblock In \emph{Proceedings of the IEEE/CVF Conference on Computer Vision
  and Pattern Recognition (CVPR)}, 2025{\natexlab{a}}.

\bibitem[Zyrianov et~al.(2025)Zyrianov, Che, Liu, and
  Wang]{zyrianov2024lidardm}
Vlas Zyrianov, Henry Che, Zhijian Liu, and Shenlong Wang.
\newblock {LidarDM}: Generative {LiDAR} simulation in a generated world.
\newblock In \emph{IEEE International Conference on Robotics and Automation
  (ICRA)}, 2025.

\bibitem[Zhou et~al.(2025{\natexlab{b}})Zhou, Jia, Zhang, Li, Zhang, Feng, Sun,
  Wong, You, and Yan]{liu2025lagen}
Sizhuo Zhou, Xiaosong Jia, Fanrui Zhang, Junjie Li, Juyong Zhang, Yukang Feng,
  Jianwen Sun, Songbur Wong, Junqi You, and Junchi Yan.
\newblock {LaGen}: Towards autoregressive {LiDAR} scene generation.
\newblock \emph{arXiv preprint arXiv:2511.21256}, 2025{\natexlab{b}}.

\bibitem[Liang et~al.(2026)Liang, Liu, Yang, Lu, Li, Kong, Zhao, and
  Ooi]{liu2025lidarcrafter}
Ao~Liang, Youquan Liu, Yu~Yang, Dongyue Lu, Linfeng Li, Lingdong Kong, Huaici
  Zhao, and Wei~Tsang Ooi.
\newblock {LiDARCrafter}: Dynamic {4D} world modeling from {LiDAR} sequences.
\newblock In \emph{Proceedings of the AAAI Conference on Artificial
  Intelligence (AAAI)}, volume~40, 2026.

\bibitem[Zyrianov et~al.(2022)Zyrianov, Zhu, and Wang]{zyrianov2022lidargen}
Vlas Zyrianov, Xiyue Zhu, and Shenlong Wang.
\newblock Learning to generate realistic {LiDAR} point clouds.
\newblock In \emph{European Conference on Computer Vision (ECCV)}, 2022.

\bibitem[Nakashima and Kurazume(2024)]{nakashima2023r2dm}
Kazuto Nakashima and Ryo Kurazume.
\newblock {LiDAR} data synthesis with denoising diffusion probabilistic models.
\newblock In \emph{IEEE International Conference on Robotics and Automation
  (ICRA)}, 2024.

\bibitem[Nakashima et~al.(2025)Nakashima, Liu, Miyawaki, Iwashita, and
  Kurazume]{nakashima2024r2flow}
Kazuto Nakashima, Xiaowen Liu, Tomoya Miyawaki, Yumi Iwashita, and Ryo
  Kurazume.
\newblock Fast {LiDAR} data generation with rectified flows.
\newblock In \emph{IEEE International Conference on Robotics and Automation
  (ICRA)}, 2025.

\bibitem[Ran et~al.(2024)Ran, Guizilini, and Wang]{ran2024lidm}
Haoxi Ran, Vitor Guizilini, and Yue Wang.
\newblock Towards realistic scene generation with {LiDAR} diffusion models.
\newblock In \emph{Proceedings of the IEEE/CVF Conference on Computer Vision
  and Pattern Recognition (CVPR)}, 2024.

\bibitem[Hu et~al.(2024)Hu, Zhang, and Hu]{hu2024rangeldm}
Qianjiang Hu, Zhimin Zhang, and Wei Hu.
\newblock {RangeLDM}: Fast realistic {LiDAR} point cloud generation.
\newblock \emph{arXiv preprint arXiv:2403.10094}, 2024.

\bibitem[Xiong et~al.(2023)Xiong, Ma, Wang, and Urtasun]{xiong2023ultralidar}
Yuwen Xiong, Wei-Chiu Ma, Jingkang Wang, and Raquel Urtasun.
\newblock Learning compact representations for {LiDAR} completion and
  generation.
\newblock In \emph{Proceedings of the IEEE/CVF Conference on Computer Vision
  and Pattern Recognition (CVPR)}, 2023.

\bibitem[Li et~al.(2025)Li, Guo, Liu, Zou, Ding, Chen, Zhu, Tan, Zhang, Wang,
  Zhou, Zhang, Qi, Zhao, Yang, Zeng, and Jin]{li2025uniscene}
Bohan Li, Jiazhe Guo, Hongsi Liu, Yingshuang Zou, Yikang Ding, Xiwu Chen,
  Hu~Zhu, Feiyang Tan, Chi Zhang, Tiancai Wang, Shuchang Zhou, Li~Zhang,
  Xiaojuan Qi, Hao Zhao, Mu~Yang, Wenjun Zeng, and Xin Jin.
\newblock {UniScene}: Unified occupancy-centric driving scene generation.
\newblock In \emph{Proceedings of the IEEE/CVF Conference on Computer Vision
  and Pattern Recognition (CVPR)}, 2025.

\bibitem[Ren et~al.(2024)Ren, Lu, Liang, Wu, Ling, Chen, Fidler, Williams, and
  Huang]{ren2024scube}
Xuanchi Ren, Yifan Lu, Hanxue Liang, Jay~Zhangjie Wu, Huan Ling, Mike Chen,
  Sanja Fidler, Francis Williams, and Jiahui Huang.
\newblock {SCube}: Instant large-scale scene reconstruction using {VoxSplats}.
\newblock In \emph{Advances in Neural Information Processing Systems
  (NeurIPS)}, 2024.

\bibitem[Xu et~al.(2026)Xu, Liang, Liu, Li, Kong, Liu, and Liu]{xu2026u4d}
Xiang Xu, Alan Liang, Youquan Liu, Linfeng Li, Lingdong Kong, Ziwei Liu, and
  Qingshan Liu.
\newblock {U4D}: Uncertainty-aware {4D} world modeling from {LiDAR} sequences.
\newblock In \emph{Proceedings of the IEEE/CVF Conference on Computer Vision
  and Pattern Recognition (CVPR)}, 2026.

\bibitem[Ren et~al.(2025)Ren, Lu, Cao, Gao, Huang, Sabour, Shen, Pfaff, Wu,
  Chen, Kim, Gao, Leal-Taix{\'e}, Chen, Fidler, and Ling]{cosmosdrivedreams}
Xuanchi Ren, Yifan Lu, Tianshi Cao, Ruiyuan Gao, Shengyu Huang, Amirmojtaba
  Sabour, Tianchang Shen, Tobias Pfaff, Jay~Zhangjie Wu, Runjian Chen,
  Seung~Wook Kim, Jun Gao, Laura Leal-Taix{\'e}, Mike Chen, Sanja Fidler, and
  Huan Ling.
\newblock {Cosmos-Drive-Dreams}: Scalable synthetic driving data generation
  with world foundation models.
\newblock arXiv preprint arXiv:2506.09042, 2025.

\bibitem[Wang et~al.(2026)Wang, Sun, Bai, Casser, Peng, Zhu, Shih, Masotto, Su,
  Parvate, Ge, Bieske, Anguelov, Tan, and Jiang]{wang2026sensor2sensor}
Jiahao Wang, Bo~Sun, Yijing Bai, Vincent Casser, Songyou Peng, Zehao Zhu,
  Meng-Li Shih, Xander Masotto, Shih-Yang Su, Kanaad~V. Parvate, Tiancheng Ge,
  Linn Bieske, Dragomir Anguelov, Mingxing Tan, and Chiyu~Max Jiang.
\newblock {Sensor2Sensor}: Cross-embodiment sensor conversion for autonomous
  driving.
\newblock In \emph{Proceedings of the IEEE/CVF Conference on Computer Vision
  and Pattern Recognition (CVPR)}, 2026.

\bibitem[Yang et~al.(2023)Yang, Chen, Wang, Manivasagam, Ma, Yang, and
  Urtasun]{yang2023unisim}
Ze~Yang, Yun Chen, Jingkang Wang, Sivabalan Manivasagam, Wei-Chiu Ma,
  Anqi~Joyce Yang, and Raquel Urtasun.
\newblock {UniSim}: A neural closed-loop sensor simulator.
\newblock In \emph{Proceedings of the IEEE/CVF Conference on Computer Vision
  and Pattern Recognition (CVPR)}, 2023.

\bibitem[Tonderski et~al.(2024)Tonderski, Lindstr{\"o}m, Hess, Ljungbergh,
  Svensson, and Petersson]{tonderski2024neurad}
Adam Tonderski, Carl Lindstr{\"o}m, Georg Hess, William Ljungbergh, Lennart
  Svensson, and Christoffer Petersson.
\newblock {NeuRAD}: Neural rendering for autonomous driving.
\newblock In \emph{Proceedings of the IEEE/CVF Conference on Computer Vision
  and Pattern Recognition (CVPR)}, 2024.

\bibitem[Tao et~al.(2023)Tao, Gao, Wang, Lao, Chen, Zhao, Hao, Liang, Salzmann,
  and Yu]{tao2023lidarnerf}
Tang Tao, Longfei Gao, Guangrun Wang, Yixing Lao, Peng Chen, Hengshuang Zhao,
  Dayang Hao, Xiaodan Liang, Mathieu Salzmann, and Kaicheng Yu.
\newblock {LiDAR-NeRF}: Novel {LiDAR} view synthesis via neural radiance
  fields.
\newblock \emph{arXiv preprint arXiv:2304.10406}, 2023.

\bibitem[Turki et~al.(2026)Turki, Wu, Kang, Martinez~Esturo, Huang, Li, Gojcic,
  and de~Lutio]{turki2026simuli}
Haithem Turki, Qi~Wu, Xin Kang, Janick Martinez~Esturo, Shengyu Huang, Ruilong
  Li, Zan Gojcic, and Riccardo de~Lutio.
\newblock {SimULi}: Real-time {LiDAR} and camera simulation with unscented
  transforms.
\newblock In \emph{International Conference on Learning Representations}, 2026.

\bibitem[Yang et~al.(2025)Yang, Huang, Chen, Wang, Li, You, Igl, Sharma,
  Karkus, Xu, Ivanovic, Wang, and Pavone]{storm}
Jiawei Yang, Jiahui Huang, Yuxiao Chen, Yan Wang, Boyi Li, Yurong You,
  Maximilian Igl, Apoorva Sharma, Peter Karkus, Danfei Xu, Boris Ivanovic, Yue
  Wang, and Marco Pavone.
\newblock Storm: Spatio-temporal reconstruction model for large-scale outdoor
  scenes.
\newblock \emph{arXiv preprint arXiv:2501.00602}, 2025.

\bibitem[Chen et~al.(2024)Chen, Wang, Yang, Manivasagam, and
  Urtasun]{chen2024g3r}
Yun Chen, Jingkang Wang, Ze~Yang, Sivabalan Manivasagam, and Raquel Urtasun.
\newblock {G3R}: Gradient guided generalizable reconstruction.
\newblock In \emph{European Conference on Computer Vision (ECCV)}, 2024.

\bibitem[Lu et~al.(2025)Lu, Xu, Zheng, Zhang, Zhan, Du, Tomizuka, Keutzer, and
  Chen]{lu2025drivingrecon}
Hao Lu, Tianshuo Xu, Wenzhao Zheng, Yunpeng Zhang, Wei Zhan, Dalong Du,
  Masayoshi Tomizuka, Kurt Keutzer, and Yingcong Chen.
\newblock {DrivingRecon}: Large {4D} gaussian reconstruction model for
  autonomous driving.
\newblock In \emph{Advances in Neural Information Processing Systems
  (NeurIPS)}, 2025.

\bibitem[Wang et~al.(2025)Wang, Che, Chen, Yang, Goli, Manivasagam, and
  Urtasun]{wang2025flux4d}
Jingkang Wang, Henry Che, Yun Chen, Ze~Yang, Lily Goli, Sivabalan Manivasagam,
  and Raquel Urtasun.
\newblock {Flux4D}: Flow-based unsupervised {4D} reconstruction.
\newblock In \emph{Advances in Neural Information Processing Systems
  (NeurIPS)}, 2025.

\bibitem[Chen et~al.(2026)Chen, Xiong, Chen, Li, Wang, Luo, Chen, Sun, Wang,
  Chen, Li, Zhang, Ye, and Zhao]{chen2026dggt}
Xiaoxue Chen, Ziyi Xiong, Yuantao Chen, Gen Li, Nan Wang, Hongcheng Luo, Long
  Chen, Haiyang Sun, Bing Wang, Guang Chen, Hongyang Li, Ya-Qin Zhang, Hangjun
  Ye, and Hao Zhao.
\newblock {DGGT}: Feedforward {4D} reconstruction of dynamic driving scenes
  using unposed images.
\newblock In \emph{Proceedings of the IEEE/CVF Conference on Computer Vision
  and Pattern Recognition (CVPR)}, 2026.

\bibitem[{NVIDIA}(2026)]{nvidia2026instantnurec}
{NVIDIA}.
\newblock Instant nurec: Feed-forward 3d gaussian reconstruction for driving
  scene simulation.
\newblock Technical report, NVIDIA, 2026.

\bibitem[Wu et~al.(2025)Wu, Zhang, Turki, Ren, Gao, Shou, Fidler, Gojcic, and
  Ling]{wu2025difix3d}
Jay~Zhangjie Wu, Yuxuan Zhang, Haithem Turki, Xuanchi Ren, Jun Gao, Mike~Zheng
  Shou, Sanja Fidler, Zan Gojcic, and Huan Ling.
\newblock {DIFIX3D+}: Improving {3D} reconstructions with single-step diffusion
  models.
\newblock In \emph{Proceedings of the IEEE/CVF Conference on Computer Vision
  and Pattern Recognition (CVPR)}, 2025.

\bibitem[Che et~al.(2026)Che, Wang, Chen, Yang, Manivasagam, and
  Urtasun]{che2026genre}
Henry Che, Jingkang Wang, Yun Chen, Ze~Yang, Sivabalan Manivasagam, and Raquel
  Urtasun.
\newblock Diffusion-guided generalizable enhancer for urban scene
  reconstruction.
\newblock In \emph{IEEE International Conference on Robotics and Automation
  (ICRA)}, 2026.

\bibitem[Schmid et~al.(2026)Schmid, von L{\"u}tzow, Hladk{\'y}, Dai, and
  Nie{\ss}ner]{schmid2026genrecon}
Katharina Schmid, Nicolas von L{\"u}tzow, Jozef Hladk{\'y}, Angela Dai, and
  Matthias Nie{\ss}ner.
\newblock {GenRecon}: Bridging generative priors for multi-view {3D} scene
  reconstruction.
\newblock \emph{arXiv preprint arXiv:2605.23888}, 2026.

\bibitem[Yin et~al.(2025)Yin, Zhang, Chang, Feng, Fan, Yang, Pun, Zhang, and
  Cun]{yin2026gsfixer}
Xingyilang Yin, Qi~Zhang, Jiahao Chang, Ying Feng, Qingnan Fan, Xi~Yang,
  Chi-Man Pun, Huaqi Zhang, and Xiaodong Cun.
\newblock {GSFixer}: Improving {3D} gaussian splatting with reference-guided
  video diffusion priors.
\newblock \emph{arXiv preprint arXiv:2508.09667}, 2025.

\bibitem[Zhu et~al.(2026)Zhu, Narayana, Stary, Hutchcroft, Wetzstein, and
  Armeni]{zhu2026gaussfusion}
Liyuan Zhu, Manjunath Narayana, Michal Stary, Will Hutchcroft, Gordon
  Wetzstein, and Iro Armeni.
\newblock {GaussFusion}: Improving {3D} reconstruction in the wild with a
  geometry-informed video generator.
\newblock In \emph{Proceedings of the IEEE/CVF Conference on Computer Vision
  and Pattern Recognition (CVPR)}, 2026.

\bibitem[Wei et~al.(2025)Wei, Leutenegger, and Schaefer]{wei2026gsfix3d}
Jiaxin Wei, Stefan Leutenegger, and Simon Schaefer.
\newblock {GSFix3D}: Diffusion-guided repair of novel views in gaussian
  splatting.
\newblock \emph{arXiv preprint arXiv:2508.14717}, 2025.

\bibitem[Khurana et~al.(2023)Khurana, Hu, Held, and Ramanan]{khurana2023point}
Tarasha Khurana, Peiyun Hu, David Held, and Deva Ramanan.
\newblock Point cloud forecasting as a proxy for {4D} occupancy forecasting.
\newblock In \emph{Proceedings of the IEEE/CVF Conference on Computer Vision
  and Pattern Recognition (CVPR)}, 2023.

\bibitem[Agro et~al.(2024)Agro, Sykora, Casas, Gilles, and
  Urtasun]{agro2024uno}
Ben Agro, Quinlan Sykora, Sergio Casas, Thomas Gilles, and Raquel Urtasun.
\newblock {UnO}: Unsupervised occupancy fields for perception and forecasting.
\newblock In \emph{Proceedings of the IEEE/CVF Conference on Computer Vision
  and Pattern Recognition (CVPR)}, 2024.

\bibitem[Diehl et~al.(2025)Diehl, Sykora, Agro, Gilles, Casas, and
  Urtasun]{diehl2025dio}
Christopher Diehl, Quinlan Sykora, Ben Agro, Thomas Gilles, Sergio Casas, and
  Raquel Urtasun.
\newblock {DIO}: Decomposable implicit {4D} occupancy-flow world model.
\newblock In \emph{Proceedings of the IEEE/CVF Conference on Computer Vision
  and Pattern Recognition (CVPR)}, 2025.

\bibitem[Yang et~al.(2024)Yang, Chen, Sun, and Li]{yang2024vidar}
Zetong Yang, Li~Chen, Yanan Sun, and Hongyang Li.
\newblock Visual point cloud forecasting enables scalable autonomous driving.
\newblock In \emph{Proceedings of the IEEE/CVF Conference on Computer Vision
  and Pattern Recognition (CVPR)}, 2024.

\bibitem[Zhang et~al.(2024)Zhang, Xiong, Yang, Casas, Hu, and
  Urtasun]{hu2024copilot4d}
Lunjun Zhang, Yuwen Xiong, Ze~Yang, Sergio Casas, Rui Hu, and Raquel Urtasun.
\newblock {Copilot4D}: Learning unsupervised world models for autonomous
  driving via discrete diffusion.
\newblock In \emph{International Conference on Learning Representations}, 2024.

\bibitem[{Team Wan} et~al.(2025){Team Wan}, Wang, Ai, Wen, Mao, Xie, Chen, Yu,
  Zhao, Yang, Zeng, Wang, Zhang, Zhou, Wang, Chen, Zhu, Zhao, Yan, Huang, Feng,
  Zhang, Li, Wu, Chu, Feng, Zhang, Sun, Fang, Wang, Gui, Weng, Shen, Lin, Wang,
  Wang, Zhou, Wang, Shen, Yu, Shi, Huang, Xu, Kou, Lv, Li, Liu, Wang, Zhang,
  Huang, Li, Wu, Liu, Pan, Zheng, Hong, Shi, Feng, Jiang, Han, Wu, and
  Liu]{wan}
{Team Wan}, Ang Wang, Baole Ai, Bin Wen, Chaojie Mao, Chen-Wei Xie, Di~Chen,
  Feiwu Yu, Haiming Zhao, Jianxiao Yang, Jianyuan Zeng, Jiayu Wang, Jingfeng
  Zhang, Jingren Zhou, Jinkai Wang, Jixuan Chen, Kai Zhu, Kang Zhao, Keyu Yan,
  Lianghua Huang, Mengyang Feng, Ningyi Zhang, Pandeng Li, Pingyu Wu, Ruihang
  Chu, Ruili Feng, Shiwei Zhang, Siyang Sun, Tao Fang, Tianxing Wang, Tianyi
  Gui, Tingyu Weng, Tong Shen, Wei Lin, Wei Wang, Wei Wang, Wenmeng Zhou, Wente
  Wang, Wenting Shen, Wenyuan Yu, Xianzhong Shi, Xiaoming Huang, Xin Xu, Yan
  Kou, Yangyu Lv, Yifei Li, Yijing Liu, Yiming Wang, Yingya Zhang, Yitong
  Huang, Yong Li, You Wu, Yu~Liu, Yulin Pan, Yun Zheng, Yuntao Hong, Yupeng
  Shi, Yutong Feng, Zeyinzi Jiang, Zhen Han, Zhi-Fan Wu, and Ziyu Liu.
\newblock {Wan}: Open and advanced large-scale video generative models.
\newblock \emph{arXiv preprint arXiv:2503.20314}, 2025.

\bibitem[Liu et~al.(2026)Liu, Zhang, Shao, and Lu]{liu2026l3dr}
Quan Liu, Xiaoqin Zhang, Ling Shao, and Shijian Lu.
\newblock {L3DR}: {3D}-aware {LiDAR} diffusion and rectification.
\newblock In \emph{Proceedings of the IEEE/CVF Conference on Computer Vision
  and Pattern Recognition (CVPR)}, 2026.

\bibitem[Hassani et~al.(2023)Hassani, Walton, Li, Li, and
  Shi]{hassani2023natten}
Ali Hassani, Steven Walton, Jiachen Li, Shen Li, and Humphrey Shi.
\newblock Neighborhood attention transformer.
\newblock In \emph{Proceedings of the IEEE/CVF Conference on Computer Vision
  and Pattern Recognition (CVPR)}, 2023.

\bibitem[Blattmann et~al.(2023)Blattmann, Rombach, Ling, Dockhorn, Kim, Fidler,
  and Kreis]{blattmann2023videoldm}
Andreas Blattmann, Robin Rombach, Huan Ling, Tim Dockhorn, Seung~Wook Kim,
  Sanja Fidler, and Karsten Kreis.
\newblock Align your latents: High-resolution video synthesis with latent
  diffusion models.
\newblock In \emph{Proceedings of the IEEE/CVF Conference on Computer Vision
  and Pattern Recognition (CVPR)}, 2023.

\bibitem[NVIDIA et~al.(2025)NVIDIA, Ali, Bai, Bala, Balaji, Blakeman, Cai, Cao,
  Cao, Cha, Chao, Chattopadhyay, Chen, Chen, Chen, Cheng, Cui, Diamond, Ding,
  Fan, Fan, Feng, Ferroni, Fidler, Fu, Gao, Ge, Gu, Gupta, Gururani, El~Hanafi,
  Hassani, Hao, Huffman, Jang, Jannaty, Kautz, Lam, Li, Li, Liao, Lin, Lin,
  Lin, Ling, Liu, Liu, Lu, Luo, Ma, Mao, Mo, Nah, Narang, Panaskar, Pavao,
  Pham, Ramezanali, Reda, Reed, Ren, Shao, Shen, Shi, Song, Stefaniak, Sun,
  Tang, Tasmeen, Tchapmi, Tseng, Varghese, Wang, Wang, Wang, Wang, Wang, Wei,
  Xu, Yang, Yang, Ye, Ye, Zeng, Zhang, Zhang, Zheng, Zhu, and Zhu]{cosmos}
NVIDIA, Arslan Ali, Junjie Bai, Maciej Bala, Yogesh Balaji, Aaron Blakeman,
  Tiffany Cai, Jiaxin Cao, Tianshi Cao, Elizabeth Cha, Yu-Wei Chao, Prithvijit
  Chattopadhyay, Mike Chen, Yongxin Chen, Yu~Chen, Shuai Cheng, Yin Cui, Jenna
  Diamond, Yifan Ding, Jiaojiao Fan, Linxi Fan, Liang Feng, Francesco Ferroni,
  Sanja Fidler, Xiao Fu, Ruiyuan Gao, Yunhao Ge, Jinwei Gu, Aryaman Gupta,
  Siddharth Gururani, Imad El~Hanafi, Ali Hassani, Zekun Hao, Jacob Huffman,
  Joel Jang, Pooya Jannaty, Jan Kautz, Grace Lam, Xuan Li, Zhaoshuo Li,
  Maosheng Liao, Chen-Hsuan Lin, Tsung-Yi Lin, Yen-Chen Lin, Huan Ling, Ming-Yu
  Liu, Xian Liu, Yifan Lu, Alice Luo, Qianli Ma, Hanzi Mao, Kaichun Mo,
  Seungjun Nah, Yashraj Narang, Abhijeet Panaskar, Lindsey Pavao, Trung Pham,
  Morteza Ramezanali, Fitsum Reda, Scott Reed, Xuanchi Ren, Haonan Shao, Yue
  Shen, Stella Shi, Shuran Song, Bartosz Stefaniak, Shangkun Sun, Shitao Tang,
  Sameena Tasmeen, Lyne Tchapmi, Wei-Cheng Tseng, Jibin Varghese, Andrew~Z.
  Wang, Hao Wang, Haoxiang Wang, Heng Wang, Ting-Chun Wang, Fangyin Wei, Jiashu
  Xu, Dinghao Yang, Xiaodong Yang, Haotian Ye, Seonghyeon Ye, Xiaohui Zeng,
  Jing Zhang, Qinsheng Zhang, Kaiwen Zheng, Andrew Zhu, and Yuke Zhu.
\newblock World simulation with video foundation models for physical ai.
\newblock \emph{arXiv preprint arXiv:2511.00062}, 2025.

\bibitem[Lipman et~al.(2023)Lipman, Chen, Ben-Hamu, Nickel, and
  Le]{lipman2022flow}
Yaron Lipman, Ricky T.~Q. Chen, Heli Ben-Hamu, Maximilian Nickel, and Matthew
  Le.
\newblock Flow matching for generative modeling.
\newblock In \emph{International Conference on Learning Representations}, 2023.

\bibitem[Liu et~al.(2023)Liu, Gong, and Liu]{liu2022flow}
Xingchao Liu, Chengyue Gong, and Qiang Liu.
\newblock Flow straight and fast: Learning to generate and transfer data with
  rectified flow.
\newblock In \emph{International Conference on Learning Representations}, 2023.

\bibitem[Bai et~al.(2025)Bai, Chen, Liu, Wang, Ge, Song, Dang, Wang, Wang,
  Tang, Zhong, Zhu, Yang, Li, Wan, Wang, Ding, Fu, Xu, Ye, Zhang, Xie, Cheng,
  Zhang, Yang, Xu, and Lin]{qwen2.5}
Shuai Bai, Keqin Chen, Xuejing Liu, Jialin Wang, Wenbin Ge, Sibo Song, Kai
  Dang, Peng Wang, Shijie Wang, Jun Tang, Humen Zhong, Yuanzhi Zhu, Mingkun
  Yang, Zhaohai Li, Jianqiang Wan, Pengfei Wang, Wei Ding, Zheren Fu, Yiheng
  Xu, Jiabo Ye, Xi~Zhang, Tianbao Xie, Zesen Cheng, Hang Zhang, Zhibo Yang,
  Haiyang Xu, and Junyang Lin.
\newblock Qwen2.5-vl technical report.
\newblock \emph{arXiv preprint arXiv:2502.13923}, 2025.

\bibitem[{NVIDIA} et~al.(2025{\natexlab{a}}){NVIDIA}, Azzolini, Brandon,
  Chattopadhyay, Chen, Chu, Cui, Diamond, Ding, Ferroni, Govindaraju, Gu,
  Gururani, El~Hanafi, Hao, Huffman, Jin, Johnson, Khan, Kurian, Lantz, Lee,
  Li, Li, Lin, Lin, Liu, Mathau, Ni, Pavao, Ping, Romero, Smelyanskiy, Song,
  Tchapmi, Wang, Wang, Wang, Wei, Xu, Xu, Yang, Yang, Zeng, and
  Zhang]{cosmosreason1}
{NVIDIA}, Alisson Azzolini, Hannah Brandon, Prithvijit Chattopadhyay, Huayu
  Chen, Jinju Chu, Yin Cui, Jenna Diamond, Yifan Ding, Francesco Ferroni, Rama
  Govindaraju, Jinwei Gu, Siddharth Gururani, Imad El~Hanafi, Zekun Hao, Jacob
  Huffman, Jingyi Jin, Brendan Johnson, Rizwan Khan, George Kurian, Elena
  Lantz, Nayeon Lee, Zhaoshuo Li, Xuan Li, Tsung-Yi Lin, Yen-Chen Lin, Ming-Yu
  Liu, Andrew Mathau, Yun Ni, Lindsey Pavao, Wei Ping, David~W. Romero, Misha
  Smelyanskiy, Shuran Song, Lyne Tchapmi, Andrew~Z. Wang, Boxin Wang, Haoxiang
  Wang, Fangyin Wei, Jiashu Xu, Yao Xu, Xiaodong Yang, Zhuolin Yang, Xiaohui
  Zeng, and Zhe Zhang.
\newblock {Cosmos-Reason1}: From physical common sense to embodied reasoning.
\newblock \emph{arXiv preprint arXiv:2503.15558}, 2025{\natexlab{a}}.

\bibitem[Xiao et~al.(2023)Xiao, Tian, Chen, Han, and
  Lewis]{xiao2024streamingllm}
Guangxuan Xiao, Yuandong Tian, Beidi Chen, Song Han, and Mike Lewis.
\newblock Efficient streaming language models with attention sinks.
\newblock \emph{arXiv preprint arXiv:2309.17453}, 2023.

\bibitem[Huang et~al.(2025)Huang, Li, He, Zhou, and
  Shechtman]{huang2025selfforcing}
Xun Huang, Zhengqi Li, Guande He, Mingyuan Zhou, and Eli Shechtman.
\newblock Self forcing: Bridging the train-test gap in autoregressive video
  diffusion.
\newblock \emph{arXiv preprint arXiv:2506.08009}, 2025.

\bibitem[Yin et~al.(2024)Yin, Gharbi, Zhang, Shechtman, Durand, Freeman, and
  Park]{yin2024dmd}
Tianwei Yin, Micha{\"e}l Gharbi, Richard Zhang, Eli Shechtman, Fr{\'e}do
  Durand, William~T. Freeman, and Taesung Park.
\newblock One-step diffusion with distribution matching distillation.
\newblock In \emph{Proceedings of the IEEE/CVF Conference on Computer Vision
  and Pattern Recognition (CVPR)}, 2024.

\bibitem[Gu et~al.(2025)Gu, Mao, and Shou]{gu2025far}
Yuchao Gu, Weijia Mao, and Mike~Zheng Shou.
\newblock Long-context autoregressive video modeling with next-frame
  prediction.
\newblock \emph{arXiv preprint arXiv:2503.19325}, 2025.

\bibitem[Xi et~al.(2026)Xi, Yang, Zhao, Li, Cai, Li, Lin, Zhang, Zhang, Li, Xu,
  Wu, Xu, Stoica, Han, and Keutzer]{xi2026quantvideogen}
Haocheng Xi, Shuo Yang, Yilong Zhao, Muyang Li, Han Cai, Xingyang Li, Yujun
  Lin, Zhuoyang Zhang, Jintao Zhang, Xiuyu Li, Zhiying Xu, Jun Wu, Chenfeng Xu,
  Ion Stoica, Song Han, and Kurt Keutzer.
\newblock Quant {VideoGen}: Auto-regressive long video generation via 2-bit
  {KV}-cache quantization.
\newblock \emph{arXiv preprint arXiv:2602.02958}, 2026.

\bibitem[Samuel et~al.(2026)Samuel, Tzachor, Levy, Green, Chechik, and
  Ben-Ari]{samuel2026fastar}
Dvir Samuel, Issar Tzachor, Matan Levy, Michael Green, Gal Chechik, and Rami
  Ben-Ari.
\newblock Fast autoregressive video diffusion and world models with temporal
  cache compression and sparse attention.
\newblock \emph{arXiv preprint arXiv:2602.01801}, 2026.

\bibitem[Guo et~al.(2026)Guo, Zhang, Wang, Liang, Ma, and
  Zhang]{guo2026memorize}
Yanjun Guo, Zhengqiang Zhang, Pengfei Wang, Xinyue Liang, Zhiyuan Ma, and Lei
  Zhang.
\newblock Memorize when needed: Decoupled memory control for spatially
  consistent long-horizon video generation.
\newblock \emph{arXiv preprint arXiv:2604.18215}, 2026.

\bibitem[Wu et~al.(2026)Wu, Elflein, Lucas, Russakovsky, Leal-Taix{\'e},
  Paschalidou, Lorraine, and O{\v{s}}ep]{wu2026worldtrace}
Xindi Wu, Sven Elflein, James Lucas, Olga Russakovsky, Laura Leal-Taix{\'e},
  Despoina Paschalidou, Jonathan Lorraine, and Aljo{\v{s}}a O{\v{s}}ep.
\newblock Addressable memory for video world models.
\newblock \emph{arXiv preprint arXiv:2608.07408}, 2026.

\bibitem[Sun et~al.(2020)Sun, Kretzschmar, Dotiwalla, Chouard, Patnaik, Tsui,
  Guo, Zhou, Chai, Caine, Vasudevan, Han, Ngiam, Zhao, Timofeev, Ettinger,
  Krivokon, Gao, Joshi, Zhang, Shlens, Chen, and Anguelov]{sun2020waymo}
Pei Sun, Henrik Kretzschmar, Xerxes Dotiwalla, Aurelien Chouard, Vijaysai
  Patnaik, Paul Tsui, James Guo, Yin Zhou, Yuning Chai, Benjamin Caine, Vijay
  Vasudevan, Wei Han, Jiquan Ngiam, Hang Zhao, Aleksei Timofeev, Scott
  Ettinger, Maxim Krivokon, Amy Gao, Aditya Joshi, Yu~Zhang, Jonathon Shlens,
  Zhifeng Chen, and Dragomir Anguelov.
\newblock Scalability in perception for autonomous driving: {Waymo Open
  Dataset}.
\newblock In \emph{Proceedings of the IEEE/CVF Conference on Computer Vision
  and Pattern Recognition (CVPR)}, 2020.

\bibitem[{NVIDIA}(2025)]{nvidiapai}
{NVIDIA}.
\newblock {NVIDIA Physical AI Autonomous Vehicles Dataset}.
\newblock
  \url{https://huggingface.co/datasets/nvidia/PhysicalAI-Autonomous-Vehicles},
  2025.

\bibitem[Caesar et~al.(2020)Caesar, Bankiti, Lang, Vora, Liong, Xu, Krishnan,
  Pan, Baldan, and Beijbom]{nuscenes}
Holger Caesar, Varun Bankiti, Alex~H. Lang, Sourabh Vora, Venice~Erin Liong,
  Qiang Xu, Anush Krishnan, Yu~Pan, Giancarlo Baldan, and Oscar Beijbom.
\newblock {nuScenes}: A multimodal dataset for autonomous driving.
\newblock In \emph{Proceedings of the IEEE/CVF Conference on Computer Vision
  and Pattern Recognition (CVPR)}, 2020.

\bibitem[{NVIDIA} et~al.(2025{\natexlab{b}}){NVIDIA}, Agarwal, Ali, Bala,
  Balaji, Barker, Cai, Chattopadhyay, Chen, Cui, Ding, Dworakowski, Fan, Fenzi,
  Ferroni, Fidler, Fox, Ge, Ge, Gu, Gururani, He, Huang, Huffman, Jannaty, Jin,
  Kim, Kl{\'a}r, Lam, Lan, Leal-Taix{\'e}, Li, Li, Lin, Lin, Ling, Liu, Liu,
  Luo, Ma, Mao, Mo, Mousavian, Nah, Niverty, Page, Paschalidou, Patel, Pavao,
  Ramezanali, Reda, Ren, Sabavat, Schmerling, Shi, Stefaniak, Tang, Tchapmi,
  Tredak, Tseng, Varghese, Wang, Wang, Wang, Wang, Wei, Wei, Wu, Xu, Yang,
  Yen-Chen, Zeng, Zeng, Zhang, Zhang, Zhang, Zhao, and
  Zolkowski]{cosmos-tokenizer}
{NVIDIA}, Niket Agarwal, Arslan Ali, Maciej Bala, Yogesh Balaji, Erik Barker,
  Tiffany Cai, Prithvijit Chattopadhyay, Yongxin Chen, Yin Cui, Yifan Ding,
  Daniel Dworakowski, Jiaojiao Fan, Michele Fenzi, Francesco Ferroni, Sanja
  Fidler, Dieter Fox, Songwei Ge, Yunhao Ge, Jinwei Gu, Siddharth Gururani,
  Ethan He, Jiahui Huang, Jacob Huffman, Pooya Jannaty, Jingyi Jin, Seung~Wook
  Kim, Gergely Kl{\'a}r, Grace Lam, Shiyi Lan, Laura Leal-Taix{\'e}, Anqi Li,
  Zhaoshuo Li, Chen-Hsuan Lin, Tsung-Yi Lin, Huan Ling, Ming-Yu Liu, Xian Liu,
  Alice Luo, Qianli Ma, Hanzi Mao, Kaichun Mo, Arsalan Mousavian, Seungjun Nah,
  Sriharsha Niverty, David Page, Despoina Paschalidou, Zeeshan Patel, Lindsey
  Pavao, Morteza Ramezanali, Fitsum Reda, Xiaowei Ren, Vasanth Rao~Naik
  Sabavat, Ed~Schmerling, Stella Shi, Bartosz Stefaniak, Shitao Tang, Lyne
  Tchapmi, Przemek Tredak, Wei-Cheng Tseng, Jibin Varghese, Hao Wang, Haoxiang
  Wang, Heng Wang, Ting-Chun Wang, Fangyin Wei, Xinyue Wei, Jay~Zhangjie Wu,
  Jiashu Xu, Wei Yang, Lin Yen-Chen, Xiaohui Zeng, Yu~Zeng, Jing Zhang,
  Qinsheng Zhang, Yuxuan Zhang, Qingqing Zhao, and Artur Zolkowski.
\newblock {Cosmos} world foundation model platform for physical {AI}.
\newblock \emph{arXiv preprint arXiv:2501.03575}, 2025{\natexlab{b}}.

\bibitem[Zhang et~al.(2026)Zhang, Wu, Yang, Fan, Li, Zhang, Huang, Wang, and
  Zhao]{zhang2026utonia}
Yujia Zhang, Xiaoyang Wu, Yunhan Yang, Xianzhe Fan, Han Li, Yuechen Zhang,
  Zehao Huang, Naiyan Wang, and Hengshuang Zhao.
\newblock {Utonia}: Toward one encoder for all point clouds.
\newblock In \emph{International Conference on Machine Learning (ICML)}, 2026.

\bibitem[Hess et~al.(2025)Hess, Lindstr{\"o}m, Fatemi, Petersson, and
  Svensson]{hess2025splatad}
Georg Hess, Carl Lindstr{\"o}m, Maryam Fatemi, Christoffer Petersson, and
  Lennart Svensson.
\newblock {SplatAD}: Real-time lidar and camera rendering with {3D} gaussian
  splatting for autonomous driving.
\newblock In \emph{Proceedings of the IEEE/CVF Conference on Computer Vision
  and Pattern Recognition (CVPR)}, 2025.

\bibitem[Xiao et~al.(2021)Xiao, Shao, Hao, Zhang, Chai, Jiao, Li, Wu, Sun,
  Jiang, Wang, and Yang]{pandaset}
Pengchuan Xiao, Zhenlei Shao, Steven Hao, Zishuo Zhang, Xiaolin Chai, Judy
  Jiao, Zesong Li, Jian Wu, Kai Sun, Kun Jiang, Yunlong Wang, and Diange Yang.
\newblock {PandaSet}: Advanced sensor suite dataset for autonomous driving.
\newblock In \emph{IEEE International Intelligent Transportation Systems
  Conference (ITSC)}, 2021.

\bibitem[Wilson et~al.(2021)Wilson, Qi, Agarwal, Lambert, Singh, Khandelwal,
  Pan, Kumar, Hartnett, Pontes, Ramanan, Carr, and Hays]{Argoverse2}
Benjamin Wilson, William Qi, Tanmay Agarwal, John Lambert, Jagjeet Singh,
  Siddhesh Khandelwal, Bowen Pan, Ratnesh Kumar, Andrew Hartnett,
  Jhony~Kaesemodel Pontes, Deva Ramanan, Peter Carr, and James Hays.
\newblock Argoverse 2: Next generation datasets for self-driving perception and
  forecasting.
\newblock In \emph{Proceedings of the Neural Information Processing Systems
  Track on Datasets and Benchmarks (NeurIPS Datasets and Benchmarks 2021)},
  2021.

\end{thebibliography}

}

\end{document}